\pdfoutput=1
\documentclass[10pt,twocolumn,letterpaper]{article}

\usepackage[pagenumbers]{cvpr} %
\usepackage[accsupp]{axessibility}

\usepackage{lipsum}
\usepackage{array}
\usepackage{times}
\usepackage{epsfig}
\usepackage{graphicx}
\usepackage{float}
\usepackage{wrapfig}
\usepackage{amsmath,amssymb,amsthm}
\usepackage{algorithm,algorithmicx,algpseudocode}
\usepackage{bm,xspace}
\usepackage{comment}
\usepackage{multirow}
\usepackage{balance}
\usepackage{url}
\usepackage{booktabs}
\usepackage{etoolbox,siunitx}
\usepackage{calc}
\usepackage{pifont,hologo}
\usepackage{color}
\usepackage{adjustbox}
\usepackage{amsmath}
\usepackage{enumitem}
\usepackage{bbding}  %
\usepackage[normalem]{ulem}  %
\usepackage{contour}
\usepackage[table]{xcolor}
\usepackage{soul}%
\usepackage{tocloft} %
\usepackage{etoc} %
\PassOptionsToPackage{dvipsnames}{xcolor}
\usepackage[pagebackref,breaklinks,colorlinks,linkcolor=link,citecolor=cite]{hyperref}
\usepackage{natbib}
\PassOptionsToPackage{square,numbers,comma,sort}{natbib}

\newcommand{\authorskip}{\hspace{2.5mm}}
\newcommand{\institutionskip}{\hspace{5.0mm}}

\definecolor{blue}{HTML}{004bb3}
\definecolor{red}{HTML}{cc1100}
\definecolor{orange}{HTML}{cc7700}
\definecolor{gray}{HTML}{efefef}
\definecolor{darkgreen}{HTML}{228B22}
\definecolor{darkgray}{HTML}{757575}

\definecolor{cite}{HTML}{3270b5}
\definecolor{link}{HTML}{b53532}
\definecolor{link}{HTML}{cc1100}
\definecolor{scratch}{HTML}{001219}
\definecolor{pretrain}{HTML}{0A9396}

\newcommand{\scratch}{\textcolor{scratch}{$\mathbf{\circ}$\,}}
\newcommand{\pretrain}{\textcolor{pretrain}{$\bullet$\,}}

\contourlength{0.8pt}

\newcommand{\prettyul}[1]{
  \uline{\phantom{#1}}%
  \llap{\contour{white}{#1}}%
}

\newcommand{\figref}[1]{Fig.~\ref{#1}}
\newcommand{\tabref}[1]{Tab.~\ref{#1}}
\newcommand{\secref}[1]{Sec.~\ref{#1}}
\renewcommand{\eqref}[1]{Eq.~\ref{#1}}

\newcommand{\marktext}[2]{\adjustbox{bgcolor=#1}{\strut #2}}

\newcolumntype{x}[1]{>{\centering\arraybackslash}p{#1}}
\newcolumntype{y}[1]{>{\raggedright\arraybackslash}p{#1}}
\newcolumntype{z}[1]{>{\raggedleft\arraybackslash}p{#1}}
\newcommand{\tablestyle}[2]{\setlength{\tabcolsep}{#1}\renewcommand{\arraystretch}{#2}\centering\footnotesize}

\DeclareMathSymbol{@}{\mathord}{letters}{"3B}

\newcommand\mypara[1]{\vspace{0mm}\noindent\textbf{#1}}

\makeatletter
\DeclareRobustCommand\onedot{\futurelet\@let@token\@onedot}
\def\@onedot{\ifx\@let@token.\else.\null\fi\xspace}

\newcommand*{\Rom}[1]{\expandafter\@slowromancap\romannumeral #1@}
\newcommand*{\rom}[1]{\expandafter\romannumeral #1}

\def\1{\bm{1}}

\def\vp{{\bm{p}}}

\DeclareMathAlphabet{\mathsfit}{\encodingdefault}{\sfdefault}{m}{sl}
\SetMathAlphabet{\mathsfit}{bold}{\encodingdefault}{\sfdefault}{bx}{n}

\let\originalleft\left
\let\originalright\right
\renewcommand{\left}{\mathopen{}\mathclose\bgroup\originalleft}
\renewcommand{\right}{\aftergroup\egroup\originalright}

\def\paperID{200}
\def\confName{CVPR}
\def\confYear{2024}

\begin{document}

\title{\textcolor{blue}{P}oint \textcolor{blue}{T}ransformer \textcolor{red}{V3}: Simpler, Faster, Stronger}

\author{Xiaoyang Wu\textsuperscript{\mdseries1,2} \authorskip 
Li Jiang\textsuperscript{3} \authorskip
Peng-Shuai Wang\textsuperscript{\mdseries4} \\ 
Zhijian Liu\textsuperscript{\mdseries5} \authorskip
Xihui Liu\textsuperscript{1} \authorskip
Yu Qiao\textsuperscript{2} \authorskip
Wanli Ouyang\textsuperscript{2} \authorskip
Tong He\textsuperscript{2*} \authorskip 
Hengshuang Zhao\textsuperscript{1*}
 \\ \\
\textsuperscript{1}HKU \institutionskip
\textsuperscript{2}SH AI Lab \institutionskip
\textsuperscript{3}CUHK(SZ) \institutionskip
\textsuperscript{4}PKU \institutionskip
\textsuperscript{5}MIT \\ 
{\tt\small \url{https://github.com/Pointcept/PointTransformerV3}}
}

\twocolumn[{%
\renewcommand\twocolumn[1][]{#1}%
\vspace{-12mm}
\maketitle
\vspace{-10mm}
\begin{center}
    \captionsetup{type=figure}
    \includegraphics[width=\linewidth]{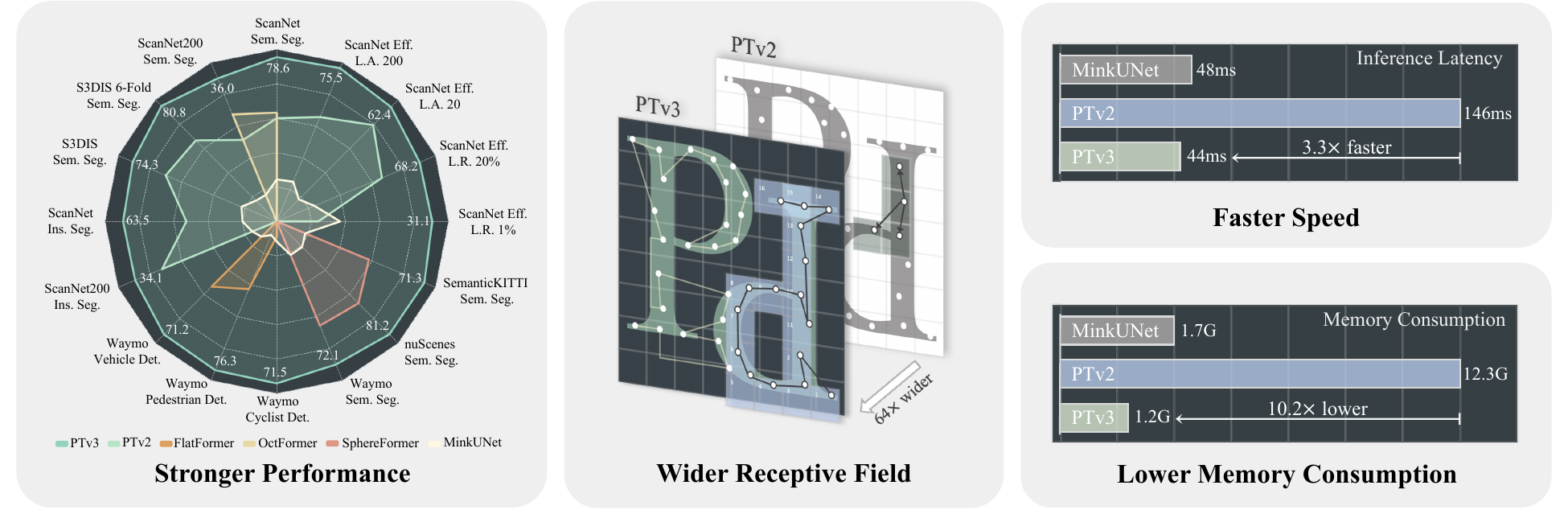}
    \vspace{-4mm}
    \captionof{figure}{\textbf{Overview of Point Transformer V3 (PTv3).} Compared to its predecessor, PTv2~\cite{wu2022point}, our PTv3 shows superiority in the following aspects: 1. \textit{Stronger performance.}  PTv3 achieves state-of-the-art results across a variety of indoor and outdoor 3D perception tasks. 2. \textit{Wider receptive field.} Benefit from the simplicity and efficiency, PTv3 expands the receptive field from 16 to 1024 points. 3. \textit{Faster speed.} PTv3 significantly increases processing speed, making it suitable for latency-sensitive applications. 4. \textit{Lower Memory Consumption.} PTv3 reduces memory usage, enhancing accessibility for broader situations.}\label{fig:teaser}
\end{center}%
}]
\begin{abstract}
\renewcommand{\thefootnote}{\fnsymbol{footnote}}
\footnotetext[1]{Corresponding author.}
\vspace{-2.5mm}
  This paper is not motivated to seek innovation within the attention mechanism. Instead, it focuses on overcoming the existing trade-offs between accuracy and efficiency within the context of point cloud processing, leveraging the power of scale. Drawing inspiration from recent advances in 3D large-scale representation learning, we recognize that model performance is more influenced by scale than by intricate design. Therefore, we present Point Transformer V3 (PTv3), which prioritizes simplicity and efficiency over the accuracy of certain mechanisms that are minor to the overall performance after scaling, such as replacing the precise neighbor search by KNN with an efficient serialized neighbor mapping of point clouds organized with specific patterns. This principle enables significant scaling, expanding the receptive field from 16 to 1024 points while remaining efficient (a 3$\times$ increase in processing speed and a 10$\times$ improvement in memory efficiency compared with its predecessor, PTv2). PTv3 attains state-of-the-art results on over 20 downstream tasks that span both indoor and outdoor scenarios. Further enhanced with multi-dataset joint training, PTv3 pushes these results to a higher level. \looseness=-1

\vspace{-4.5mm}
\end{abstract}

\section{Introduction}
\label{sec:intro}

Deep learning models have experienced rapid advancements in various areas, such as 2D vision~\cite{tian2021divide,goyal2021self,wang2022painter,kirillov2023segany} and natural language processing (NLP)~\cite{kaplan2020scaling,aribandi2022ext,touvron2023llama,openai2023gpt4}, with their progress often attributed to the effective utilization of scale, encompassing factors such as the size of datasets, the number of model parameters, the range of effective receptive field, and the computing power allocated for training.

However, in contrast to the progress made in 2D vision or NLP, the development of 3D backbones~\cite{qi2017pointnet, dai20183dmv, li2018pointcnn, wu2019pointconv} has been hindered in terms of scale, primarily due to the limited size and diversity of point cloud data available in separate domains~\cite{wu2023ppt}. Consequently, there exists a gap in applying scaling principles that have driven advancements in other fields~\cite{kaplan2020scaling}. This absence of scale often leads to a limited trade-off between accuracy and speed on 3D backbones, particularly for models based on the transformer architecture~\cite{zhao2021point, guo2021pct}. Typically, this trade-off involves sacrificing efficiency for accuracy. Such limited efficiency impedes some of these models' capacity to fully leverage the inherent strength of transformers in scaling the range of receptive fields, hindering their full potential in 3D data processing. \looseness=-1

A recent advancement~\cite{wu2023ppt} in 3D representation learning has made progress in overcoming the data scale limitation in point cloud processing by introducing a synergistic training approach spanning multiple 3D datasets. Coupled with this strategy, the efficient convolutional backbone~\cite{choy20194d} has effectively bridged the accuracy gap commonly associated with point cloud transformers~\cite{wu2022point, lai2022stratified}. However, point cloud transformers themselves have not yet fully benefited from this privilege of scale due to their efficiency gap compared to sparse convolution. This discovery shapes the initial motivation for our work:  \textit{to re-weigh the design choices in point transformers, with the lens of the scaling principle}. We posit that model performance is more significantly influenced by scale than by intricate design.

Therefore, we introduce Point Transformer V3 (PTv3), which prioritizes simplicity and efficiency over the accuracy of certain mechanisms, thereby enabling scalability. Such adjustments have an ignorable impact on overall performance after scaling. Specifically, PTv3 makes the following adaptations to achieve superior efficiency and scalability:\looseness=-1
\begin{itemize}[leftmargin=4mm, itemsep=0mm, topsep=0mm, partopsep=0mm]
    \item Inspired by two recent advancements~\cite{Wang2023OctFormer, liu2023flatformer} and recognizing the scalability benefits of structuring unstructured point clouds, PTv3 shifts from the traditional spatial proximity defined by K-Nearest Neighbors (KNN) query, accounting for 28\% of the forward time. Instead, it explores the potential of serialized neighborhoods in point clouds, organized according to specific patterns.
    \item PTv3 replaces more complex attention patch interaction mechanisms, like shift-window (impeding the fusion of attention operators) and the neighborhood mechanism (causing high memory consumption), with a streamlined approach tailored for serialized point clouds.
    \item PTv3 eliminates the reliance on relative positional encoding, which accounts for 26\% of the forward time, in favor of a simpler prepositive sparse convolutional layer.
\end{itemize}
We consider these designs as intuitive choices driven by the scaling principles and advancements in existing point cloud transformers. Importantly, this paper underscores the critical importance of recognizing how scalability affects backbone design, instead of detailed module designs.

This principle significantly enhances scalability, overcoming traditional trade-offs between accuracy and efficiency (see \figref{fig:teaser}). PTv3, compared to its predecessor, has achieved a 3.3$\times$ increase in inference speed and a 10.2$\times$ reduction in memory usage. More importantly, PTv3 capitalizes on its inherent ability to scale the range of perception, expanding its receptive field from 16 to 1024 points while maintaining efficiency. This scalability underpins its superior performance in real-world perception tasks, where PTv3 achieves state-of-the-art results across over 20 downstream tasks in both indoor and outdoor scenarios. Further augmenting its data scale with multi-dataset training~\cite{wu2023ppt}, PTv3 elevates these results even more. We hope that our insights will inspire future research in this direction.

\section{Related Work}
\label{sec:related}

\mypara{3D understanding.}
Conventionally, deep neural architectures for understanding 3D point cloud data can be broadly classified into three categories based on their approach to modeling point clouds: projection-based, voxel-based, and point-based methods. Projection-based methods project 3D points onto various image planes and utilize 2D CNN-based backbones for feature extraction~\cite{su15mvcnn,li2016vehicle,chen2017multi,lang2019pointpillars}. Voxel-based approaches transform point clouds into regular voxel grids to facilitate 3D convolution operations~\cite{maturana2015voxnet,song2017semantic}, with their efficiency subsequently enhanced by sparse convolution~\cite{wang2017ocnn,graham20183d,choy20194d}. 
However, they often lack scalability in terms of the kernel sizes. Point-based methods, by contrast, process point clouds directly~\cite{qi2017pointnet,qi2017pointnet++,zhao2019pointweb,thomas2019kpconv, ma2022rethinking} and have recently seen a shift towards transformer-based architectures~\cite{guo2021pct,zhao2021point,wu2022point, robert2023spt, yang2023swin3d}.
While these methods are powerful, their efficiency is frequently constrained by the unstructured nature of point clouds, which poses challenges to scaling their designs.

\mypara{Serialization-based method.} 
Recent works~\cite{Wang2023OctFormer, liu2023flatformer, chen2022efficient} have introduced approaches diverging from the traditional paradigms of point cloud processing, which we categorized as serialization-based. These methods structure point clouds by sorting them according to specific patterns, transforming unstructured, irregular point clouds into manageable sequences while preserving certain spatial proximity. OctFormer~\cite{Wang2023OctFormer} inherits order during octreelization, akin to z-order, offering scalability but still constrained by the octree structure itself. FlatFormer~\cite{liu2023flatformer}, on the other hand, employs a window-based sorting strategy for grouping point pillars, akin to window partitioning. However, this design lacks scalability in the receptive field and is more suited to pillar-based 3D object detectors. These pioneering works mark the inception of serialization-based methods. Our PTv3 builds on this foundation, defining and exploring the full potential of point cloud serialization.

\mypara{3D representation learning.}
In contrast to 2D domains, where large-scale pre-training has become a standard approach for enhancing downstream tasks~\cite{caron2021emerging}, 3D representation learning is still in a phase of exploration. Most studies still rely on training models from scratch using specific target datasets~\cite{xie2020pointcontrast}. While major efforts in 3D representation learning focused on individual objects~\cite{wang2019deep,sauder2019self,sanghi2020info3d, Pang2022pointmae, Yu2022pointbert}, some recent advancements have redirected attention towards training on real-world scene-centric point clouds~\cite{xie2020pointcontrast,hou2021exploring,wu2023masked, jiang2023msp, zhu2023ponderv2}. This shift signifies a major step forward in 3D scene understanding. Notably, Point Prompt Training (PPT)~\cite{wu2023ppt} introduces a new paradigm for large-scale representation learning through multi-dataset synergistic learning, emphasizing the importance of scale. This approach greatly influences our design philosophy and initial motivation for developing PTv3, and we have incorporated this strategy in our final results.

\section{Design Principle and Pilot Study}
\label{sec:principles}
\vspace{-1mm}
In this section, we introduce the scaling principle and pilot study, which guide the design of our model.

\begin{table}[t]
    \begin{minipage}{0.48\textwidth}
    \centering
        \vspace{-4.1mm}
        \tablestyle{3.2pt}{0.95}
        \begin{tabular}{l|rrrrrr}\toprule
\multicolumn{2}{l}{Outdoor Efficiency (nuScenes)} &\multicolumn{2}{c}{Training} &\multicolumn{2}{c}{Inference} \\ \cmidrule(lr){3-4} \cmidrule(lr){5-6}
Methods &Params. &Latency &Memory &Latency &Memory \\
\midrule
MinkUNet / 3~\cite{choy20194d} &37.9M &163ms &3.3G &48ms &1.7G \\
MinkUNet / 5~\cite{choy20194d} &170.3M &455ms &5.6G &145ms &2.1G \\
MinkUNet / 7~\cite{choy20194d} &465.0M &\textbf{1120ms} &12.4G &\textbf{337ms} &2.8G \\
\midrule
PTv2 / 16~\cite{wu2022point} &12.8M &213ms &10.3G &146ms &12.3G \\
PTv2 / 24~\cite{wu2022point} &12.8M &308ms &17.6G &180ms &15.2G \\
PTv2 / 32~\cite{wu2022point} &12.8M &354ms &\textbf{21.5G} &213ms &\textbf{19.4G} \\
\midrule
\cellcolor[HTML]{efefef}PTv3 / 256~(ours) &\cellcolor[HTML]{efefef}46.2M &\cellcolor[HTML]{efefef}120ms &\cellcolor[HTML]{efefef}3.3G &\cellcolor[HTML]{efefef}44ms &\cellcolor[HTML]{efefef}1.2G \\
\cellcolor[HTML]{efefef}PTv3 / 1024~(ours) &\cellcolor[HTML]{efefef}46.2M &\cellcolor[HTML]{efefef}119ms &\cellcolor[HTML]{efefef}3.3G &\cellcolor[HTML]{efefef}44ms &\cellcolor[HTML]{efefef}1.2G \\
\cellcolor[HTML]{efefef}PTv3 / 4096~(ours) &\cellcolor[HTML]{efefef}46.2M &\cellcolor[HTML]{efefef}125ms &\cellcolor[HTML]{efefef}3.3G &\cellcolor[HTML]{efefef}45ms &\cellcolor[HTML]{efefef}1.2G \\
\bottomrule
\end{tabular}
        \vspace{-2.2mm}
        \caption{\textbf{Model efficiency.} We benchmark the training and inference efficiency of backbones with various scales of receptive field. The batch size is fixed to 1, and the number after ``/" denotes the kernel size of sparse convolution and patch size\protect\footnotemark of attention.}\label{tab:outdoor_model_efficiency}
        \vspace{-6mm}
    \end{minipage}
\end{table}

\footnotetext{\textit{Patch size} refers to the number of neighboring points considered together for self-attention mechanisms.}

\mypara{Scaling principle.} 
Conventionally, the relationship between accuracy and efficiency in model performance is characterized as a ``trade-off'', with a typical preference for accuracy at the expense of efficiency. In pursuit of this, numerous methods have been proposed with cumbersome operations. Point Transformers ~\cite{zhao2021point, wu2022point} prioritize accuracy and stability by substituting matrix multiplication in the computation of attention weights with learnable layers and normalization, potentially compromising efficiency. Similarly, Stratified Transformer~\cite{lai2022stratified} and Swin3D~\cite{yang2023swin3d} achieve improved accuracy by incorporating more complex forms of relative positional encoding, yet this often results in decreased computational speed.

Yet, the perceived trade-off between accuracy and efficiency is not absolute, with a notable counterexample emerging through the engagement with scaling strategies. Specifically, Sparse Convolution, known for its speed and memory efficiency, remains preferred in 3D large-scale pre-training. Utilizing multi-dataset joint training strategies~\cite{wu2023ppt}, Sparse Convolution~\cite{graham20183d, choy20194d} has shown significant performance improvements, increasing mIoU on ScanNet semantic segmentation from 72.2\% to 77.0\%~\cite{zhu2023ponderv2}. This outperforms PTv2 when trained from scratch by 1.6\%, all while retaining superior efficiency. 
However, such advancements have not been fully extended to point transformers, primarily due to their efficiency limitations, which present burdens in model training especially when the computing resource is constrained.

This observation leads us to hypothesize that model performance may be more significantly influenced by scale than by complex design details. We consider the possibility of trading the accuracy of certain mechanisms for simplicity and efficiency, thereby enabling scalability. By leveraging the strength of scale, such sacrifices could have an ignorable impact on overall performance. This concept forms the basis of our scaling principle for backbone design, and we practice it with our design.

\begin{figure}[t]
    \vspace{-4.1mm}
    \centering
    \includegraphics[width=\linewidth]{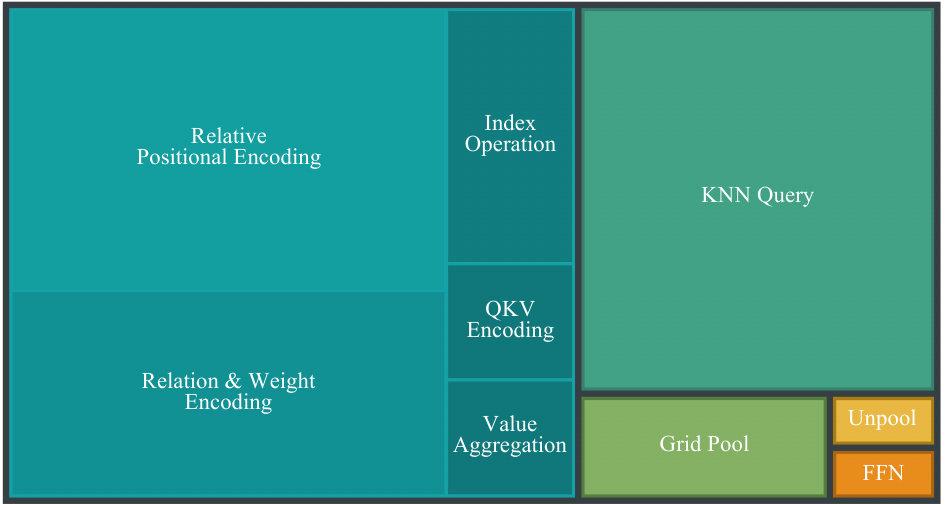}
    \vspace{-6mm}
    \caption{\textbf{Latency treemap of each components of PTv2.} We benchmark and visualize the proportion of the forward time of each component of PTv2. KNN Query and RPE occupy a total of 54\% of forward time.}
    \label{fig:ptv2_latency}
    \vspace{-7.4mm}
\end{figure}

\mypara{Breaking the curse of permutation invariance.}
Despite the demonstrated efficiency of sparse convolution, the question arises about the need for a scalable point transformer.
While multi-dataset joint training allows for data scaling and the incorporation of more layers and channels contributes to model scaling, efficiently expanding the receptive field to enhance generalization capabilities remains a challenge for convolutional backbones (refer to~\tabref{tab:outdoor_model_efficiency}). It is attention, an operator that is naturally adaptive to kernel shape, potentially to be universal.

However, current point transformers encounter challenges in scaling when adhering to the request of permutation invariance, stemming from the unstructured nature of point cloud data. In PTv1, the application of the K-Nearest Neighbors (KNN) algorithm to formulate local structures introduced computational complexities. PTv2 attempted to relieve this by halving the usage of KNN compared to PTv1. Despite this improvement, KNN still constitutes a significant computational burden, consuming 28\% of the forward time (refer to~\figref{fig:ptv2_latency}). Additionally, while Image Relative Positional Encoding (RPE) benefits from a grid layout that allows for the predefinition of relative positions, point cloud RPE must resort to computing pairwise Euclidean distances and employ learned layers or lookup tables for mapping such distances to embeddings, proves to be another source of inefficiency, occupying 26\% of the forward time (see~\figref{fig:ptv2_latency}). These extremely inefficient operations bring difficulties when scaling up the backbone.

\begin{figure*}[t]
    \vspace{-1.5mm}
    \centering
    \includegraphics[width=\linewidth]{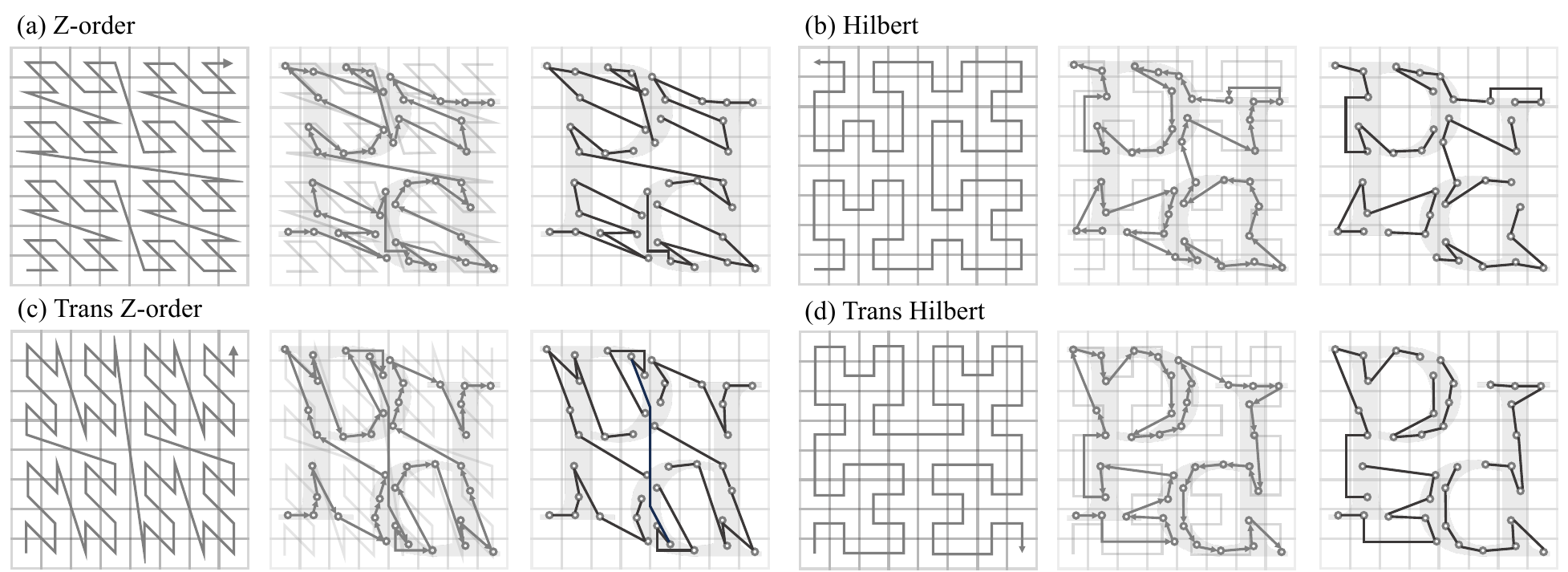}
    \vspace{-6.5mm}
    \caption{\textbf{Point cloud serialization.} We show the four patterns of serialization with a triplet visualization. For each triplet, we show the space-filling curve for serialization (left), point cloud serialization var sorting order within the space-filling curve (middle), and grouped patches of the serialized point cloud for local attention (right). Shifting across the four serialization patterns allows the attention mechanism to capture various spatial relationships and contexts, leading to an improvement in model accuracy and generalization capacity.}
    \label{fig:serialization}
    \vspace{-3mm}
\end{figure*}

Inspired by two recent advancements~\cite{Wang2023OctFormer,liu2023flatformer}, we move away from the traditional paradigm, which treats point clouds as unordered sets. Instead, we choose to ``break''  the constraints of permutation invariance by serializing point clouds into a structured format. This strategic transformation enables our method to leverage the benefits of structured data inefficiency with a compromise of the accuracy of locality-preserving property. We consider this trade-off as an entry point of our design.

\section{Point Transformer V3}
\label{sec:method}
\vspace{-2mm}

In this section, we present our designs of Point Transformer V3 (PTv3), guided by the scaling principle discussed in \secref{sec:principles}. Our approach emphasizes simplicity and speed, facilitating scalability and thereby making it stronger.

\subsection{Point Cloud Serialization}
\label{sec:serialization}
To trade the simplicity and efficiency nature of structured data, we introduce point cloud serialization, transforming unstructured point clouds into a structured format.

\mypara{Space-filling curves.} Space-filling curves~\cite{peano1990courbe} are paths that pass through every point within a higher-dimensional discrete space and preserve spatial proximity to a certain extent.
Mathematically, it can be defined as a bijective function $\varphi:  \mathbb{Z} \mapsto \mathbb{Z}^n$, where n is the dimensionality of the space, which is 3 within the context of point clouds and also can extend to a higher dimension. Our method centers on two representative space-filling curves: the z-order curve~\cite{morton1966computer} and the Hilbert curve~\cite{hilbert1935stetige}. The Z-order curve (see \figref{fig:serialization}\textcolor{link}{a}) is valued for its simplicity and ease of computation, whereas the Hilbert curve (see \figref{fig:serialization}\textcolor{link}{b}) is known for its superior locality-preserving properties compared with Z-order curve. \looseness=-1

Standard space-filling curves process the 3D space by following a sequential traversal along the x, y, and z axes, respectively. By altering the order of traversal, such as prioritizing the y-axis before the x-axis, we introduce reordered variants of standard space-filling curves. To differentiate between the standard configurations and the alternative variants of space-filling curves, we denote the latter with the prefix ``trans'’, resulting in names such as Trans Z-order (see \figref{fig:serialization}\textcolor{link}{c}) and Trans Hilbert (see \figref{fig:serialization}\textcolor{link}{d}). These variants can offer alternative perspectives on spatial relationships, potentially capturing special local relationships that the standard curve may overlook. 

\mypara{Serialized encoding.} To leverage the locality-preserving properties of space-filling curves, we employ serialized encoding, a strategy that converts a point's position into an integer reflecting its order within a given space-filling curve. Due to the bijective nature of these curves, there exists an inverse mapping $\varphi^{-1}:  \mathbb{Z}^n \mapsto \mathbb{Z}$ which allows for the transformation of a point's position $\vp_i \in \mathbb{R}^3$ into a serialization code. By projecting the point's position onto a discrete space with a grid size of $g \in \mathbb{R}$, we obtain this code as $\varphi^{-1}(\lfloor\ \vp\ /\ g\ \rfloor)$. This encoding is also adaptable to batched point cloud data. By assigning each point a 64-bit integer to record serialization code, we allocate the trailing $k$ bits to the position encoded by $\varphi^{-1}$ and the remaining leading bits to the batch index $b \in \mathbb{Z}$. Sorting the points according to this serialization code makes the batched point clouds ordered with the chosen space-filling curve pattern within each batch. The whole process can be written as follows:
\vspace{-5mm}
\begin{align*}
    \texttt{Encode}(\vp, b, g) = (b \ll k) \texttt{|} \varphi^{-1}(\lfloor\ \vp\ /\ g\ \rfloor) ,
    \label{eq:encode}
\end{align*}
 where $\ll$ denotes left bit-shift and $\texttt{|}$ denotes bitwise OR.
 
\mypara{Serialization.} As illustrated in the \textit{middle} part of triplets in \figref{fig:serialization}, the serialization of point clouds is accomplished by sorting the codes resulting from the serialized encoding. The ordering effectively rearranges the points in a manner that respects the spatial ordering defined by the given space-filling curve, which means that neighbor points in the data structure are also likely to be close in space.

In our implementation, we do not physically re-order the point clouds, but rather, we record the mappings generated by the serialization process.
This strategy maintains compatibility with various serialization patterns and provides the flexibility to transition between them efficiently. 

\vspace{-1mm}
\subsection{Serialized Attention}
\label{sec:serilization}
\vspace{-1mm}

\begin{figure}[t]
    \vspace{-1mm}
    \centering
    \includegraphics[width=\linewidth]{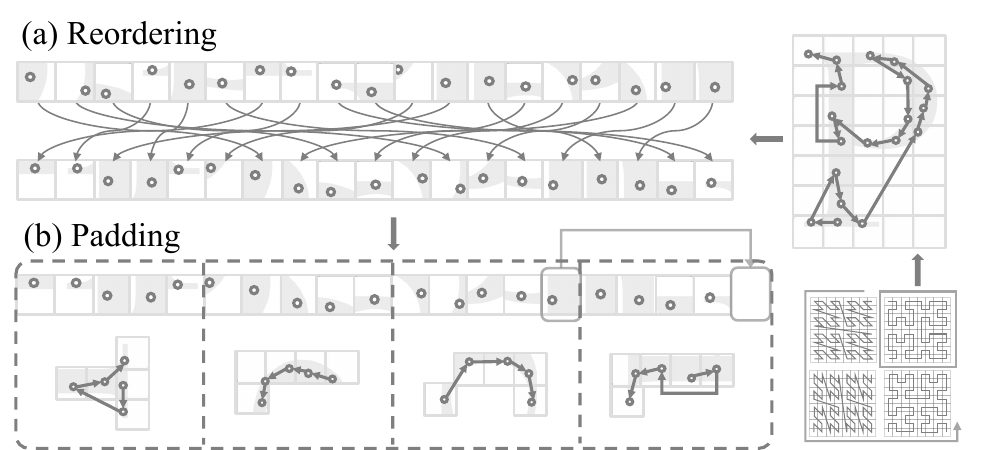}
    \vspace{-7mm}
    \caption{\textbf{Patch grouping.} (a) Reordering point cloud according to order derived from a specific serialization pattern. (b) Padding point cloud sequence by borrowing points from neighboring patches to ensure it is divisible by the designated patch size.}
    \label{fig:patch_grouping}
    \vspace{-7mm}
\end{figure}

\mypara{Re-weigh options of attention mechanism.} 
Image transformers~\cite{liu2021Swin, liu2021SwinV2, dong2022cswin}, benefiting from the structured and regular grid of pixel data, naturally prefer window~\cite{liu2021Swin} and dot-product~\cite{vaswani2017attention, dosovitskiy2020image} attention mechanisms. These methods take advantage of the fixed spatial relationships inherent to image data, allowing for efficient and scalable localized processing. However, this advantage vanishes when confronting the unstructured nature of point clouds. To adapt, previous point transformers~\cite{zhao2021point, wu2022point} introduce neighborhood attention to construct even-size attention kernels and adopt vector attention to improve model convergence on point cloud data with a more complex spatial relation.

In light of the structured nature of serialized point clouds, we choose to revisit and adopt the efficient window and dot-product attention mechanisms as our foundational approach. 
While the serialization strategy may temporarily yield a lower performance than some neighborhood construction strategies like KNN due to a reduction in precise spatial neighbor relationships,
we will demonstrate that any initial accuracy gaps can be effectively bridged by harnessing the scalability potential inherent in serialization.

Evolving from window attention, we define \textit{patch attention}, a mechanism that groups points into non-overlapping patches and performs attention within each individual patch. The effectiveness of patch attention relies on two major designs: patch grouping and patch interaction.

\mypara{Patch grouping.} Grouping points into patches within serialized point clouds has been well-explored in recent advancements~\cite{Wang2023OctFormer, liu2023flatformer}. This process is both natural and efficient, involving the simple grouping of points along the serialized order after padding. Our design for patch attention is also predicated on this strategy as presented in \figref{fig:patch_grouping}. In practice, the processes of reordering and patch padding can be integrated into a single indexing operation.

Furthermore, we illustrate patch grouping patterns derived from the four serialization patterns on the right part of triplets in \figref{fig:serialization}. This grouping strategy, in tandem with our serialization patterns, is designed to effectively broaden the attention mechanism's receptive field in the 3D space as the patch size increases while still preserving spatial neighbor relationships to a feasible degree. Although this approach may sacrifice some neighbor search accuracy when compared to KNN, the trade-off is beneficial. Given the attention's re-weighting capacity to reference points, the gains in efficiency and scalability far outweigh the minor loss in neighborhood precision (scaling it up is all we need).

\mypara{Patch interaction.} The interaction between points from different patches is critical for the model to integrate information across the entire 
point cloud. This design element counters the limitations of a non-overlapping architecture and is pivotal in making patch attention functional. Building on this insight, we investigate various designs for patch interaction as outlined below (also visualized in \figref{fig:patch_interaction}):

\begin{figure}[t]
    \vspace{-1mm}
    \centering
    \includegraphics[width=\linewidth]{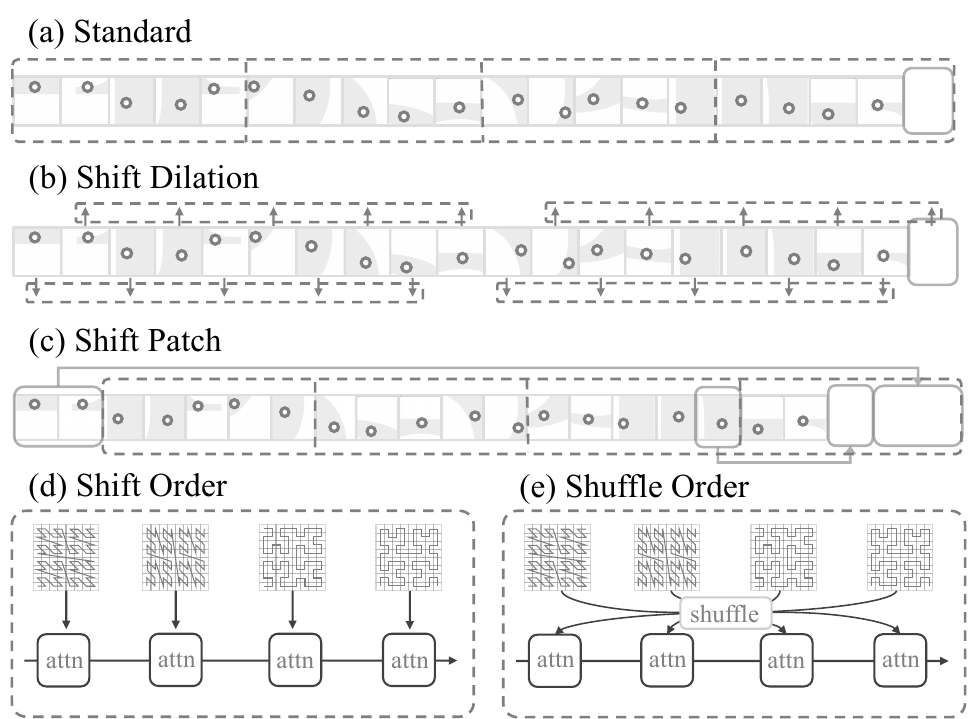}
    \vspace{-7mm}
    \caption{\textbf{Patch interaction.} (a) Standard patch grouping with a regular, non-shifted arrangement; (b) Shift Dilation where points are grouped at regular intervals, creating a dilated effect; (c) Shift Patch, which applies a shifting mechanism similar to the shift window approach; (d) Shift Order where different serialization patterns are cyclically assigned to successive attention layers; (d) Shuffle Order, where the sequence of serialization patterns is randomized before being fed to attention layers.}
    \label{fig:patch_interaction}
    \vspace{-7mm}
\end{figure}

\begin{itemize}[leftmargin=5mm, itemsep=0mm, topsep=0mm, partopsep=0mm]
    \item In \textit{Shift Dilation}~\cite{Wang2023OctFormer}, patch grouping is staggered by a specific step across the serialized point cloud, effectively extending the model's receptive field beyond the immediate neighboring points.
    \item In \textit{Shift Patch}, the positions of patches are shifted across the serialized point cloud, drawing inspiration from the shift-window strategy in image transformers~\cite{liu2021Swin}. This method maximizes the interaction among patches.
    \item In \textit{Shift Order}, the serialized order of the point cloud data is dynamically varied between attention blocks. This technique, which aligns seamlessly with our point cloud serialization method, serves to prevent the model from overfitting to a single pattern and promotes a more robust integration of features across the data.
    \item \textit{Shuffle Order$^*$}, building upon \textit{Shift Order}, introduces a random shuffle to the permutations of serialized orders. This method ensures that the receptive field of each attention layer is not limited to a single pattern, thus further enhancing the model's ability to generalize.  
\end{itemize}
We mark our main proposal with $*$ and underscore its superior performance in model ablation. 

\mypara{Positional encoding.}
To handle the voluminous data, point cloud transformers commonly employ local attention, which is reliant on relative positional encoding methods~\cite{zhao2021point, lai2022stratified, yang2023swin3d} for optimal performance. However, our observations indicate that RPEs are notably inefficient and complex. As a more efficient alternative, conditional positional encoding (CPE)~\cite{chu2021cpe, Wang2023OctFormer} is introduced for point cloud transformers, where implemented by octree-based depthwise convolutions~\cite{wang2017ocnn}. We consider this replacement to be elegant, as the implementation of RPE in point cloud transformers can essentially be regarded as a variant of large-kernel sparse convolution. Even so, a single CPE is not sufficient for the peak performance (there remains potential for an additional 0.5\% improvement when coupled with RPE). Therefore, we present an enhanced conditional positional encoding (xCPE), implemented by directly prepending a sparse convolution layer with a skip connection before the attention layer. Our experimental results demonstrate that xCPE fully unleashes the performance with a slight increase in latency of a few milliseconds compared to the standard CPE, the performance gains justify this minor trade-off.

\begin{figure}[t]
    \vspace{-1.1mm}
    \centering
    \includegraphics[width=\linewidth]{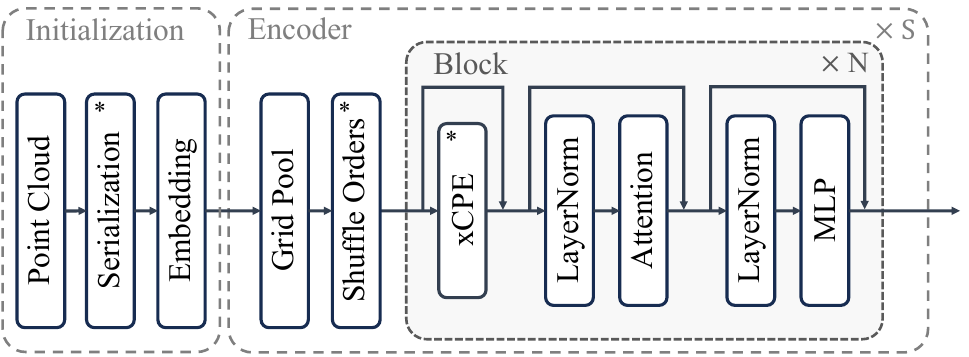}
    \vspace{-7mm}
    \caption{\textbf{Overall architecture.}}
    \label{fig:architecture}
    \vspace{-7mm}
\end{figure}

\subsection{Network Details}
\label{subsec:network_details}
In this section, we detail the macro designs of PTv3, including block structure, pooling strategy, and model architecture (visualized in \figref{fig:architecture}). Our options for these components are empirical yet also crucial to overall simplicity. Detailed ablations of these choices are available in the\prettyul{Appendix}.

\mypara{Block structure.}
We simplify the traditional block structure,  typically an extensive stack of normalization and activation layers, by adopting a pre-norm~\cite{child2019generating} structure, evaluated against the post-norm~\cite{vaswani2017attention} alternative. Additionally, we shift from Batch Normalization (BN) to Layer Normalization (LN). The proposed xCPE is prepended directly before the attention layer with a skip connection.

\mypara{Pooling strategy.}
We keep adopting the Grid Pooling introduced in PTv2, recognizing its simplicity and efficiency. Our experiments indicate that BN is essential and cannot be effectively replaced by LN. We hypothesize that BN is crucial for stabilizing the data distribution in point clouds during pooling. Additionally, the proposed Shuffle Order, with shuffle the permutation of serialized orders for Shift Order, is integrated into the pooling.

\mypara{Model architecture.} 
The architecture of PTv3 remains consistent with the U-Net~\cite{ronneberger2015unet} framework. It consists of four stage encoders and decoders, with respective block depths of [2, 2, 6, 2] and [1, 1, 1, 1]. For these stages, the grid size multipliers are set at [$\times$2, $\times$2, $\times$2, $\times$2], indicating the expansion ratio relative to the preceding pooling stage.

\section{Experiments}
\label{sec:exp}

\subsection{Main Properties}
\label{sec:ablation}
We perform an ablation study on PTv3, focusing on module design and scalability. We report the performance using the \prettyul{mean} results from the ScanNet semantic segmentation validation and measure the latencies using the \prettyul{average} values obtained from the full ScanNet validation set (with a batch size of 1) on a single RTX 4090. If not specified, default settings are presented in \prettyul{Appendix}. All of our designs are enabled by default. Default settings are marked in \marktext{gray}{gray}.

\begin{table}[t]
    \begin{minipage}{0.48\textwidth}
    \centering
        \tablestyle{1.5pt}{1.05}
        \begin{tabular}{y{22mm}|x{14mm}x{14mm}x{14mm}x{14mm}}
\toprule
Patterns & S.O. & + S.D. & + S.P. &\cellcolor[HTML]{efefef}+ Shuffle O. \\
\midrule
Z &74.3$\,_\text{\textcolor{darkgray}{54ms}}$ &75.5$\,_\text{\textcolor{darkgray}{89ms}}$ &75.8$\,_\text{\textcolor{darkgray}{86ms}}$ &\cellcolor[HTML]{efefef}74.3$\,_\text{\textcolor{darkgray}{54ms}}$ \\
Z + TZ &76.0$\,_\text{\textcolor{darkgray}{55ms}}$ &76.3$\,_\text{\textcolor{darkgray}{92ms}}$ &76.1$\,_\text{\textcolor{darkgray}{89ms}}$ &\cellcolor[HTML]{efefef}76.9$\,_\text{\textcolor{darkgray}{55ms}}$ \\
H + TH &76.2$\,_\text{\textcolor{darkgray}{60ms}}$ &76.1$\,_\text{\textcolor{darkgray}{98ms}}$ &76.2$\,_\text{\textcolor{darkgray}{94ms}}$ &\cellcolor[HTML]{efefef}76.8$\,_\text{\textcolor{darkgray}{60ms}}$ \\
\cellcolor[HTML]{efefef}Z + TZ + H + TH &\cellcolor[HTML]{efefef}76.5$\,_\text{\textcolor{darkgray}{61ms}}$ &\cellcolor[HTML]{efefef}76.8$\,_\text{\textcolor{darkgray}{99ms}}$ &\cellcolor[HTML]{efefef}76.6$\,_\text{\textcolor{darkgray}{97ms}}$ &\cellcolor[HTML]{efefef}\textbf{77.3}$\,_\text{\textcolor{darkgray}{61ms}}$ \\
\bottomrule
\end{tabular}
        \vspace{-2.5mm}
        \caption{\textbf{Serialization patterns and patch interaction.}  The first column indicates serialization patterns: Z for Z-order, TZ for Trans Z-order, H for Hilbert, and TH for Trans Hilbert. In the first row, S.O. represents Shift Order, which is the default setting also applied to other interaction strategies. S.D. stands for Shift Dilation, and S.P. signifies Shift Patch.}\label{tab:ablation_serialization_interaction}
        \vspace{2mm}
    \end{minipage} \\
    \begin{minipage}{0.48\textwidth}
    \centering
        \tablestyle{1.5pt}{1.05}
        \begin{tabular}{y{16mm}|x{12mm}x{12mm}x{12mm}x{12mm}x{12mm}}
\toprule
PE &APE &RPE &cRPE &CPE &\cellcolor[HTML]{efefef}xCPE \\\midrule
Perf. (\%) &72.1$\,_\text{\textcolor{darkgray}{50ms}}$ &75.9$\,_\text{\textcolor{darkgray}{72ms}}$ &76.8$\,_\text{\textcolor{darkgray}{101ms}}$ &76.6$\,_\text{\textcolor{darkgray}{58ms}}$ &\cellcolor[HTML]{efefef}\textbf{77.3}$\,_\text{\textcolor{darkgray}{61ms}}$ \\
\bottomrule
\end{tabular}

        \vspace{-2.5mm}
        \caption{\textbf{Positional encoding.} We compare the proposed CPE+ with APE, RPE, cRPE, and CPE. RPE and CPE are discussed in OctFormer~\cite{Wang2023OctFormer}, while cRPE is deployed by Swin3D~\cite{yang2023swin3d}.}\label{tab:ablation_positional_encoding}
        \vspace{2mm}
    \end{minipage} \\
    \begin{minipage}{0.48\textwidth}
    \centering
        \tablestyle{2pt}{1.05}
        \begin{tabular}{y{12mm}|x{8mm}x{8mm}x{8mm}x{8mm}x{8mm}x{8mm}x{8mm}}
\toprule
P.S. &16 &32 &64 &128 &256 &\cellcolor[HTML]{efefef}1024 &4096 \\\midrule
Perf. (\%) &75.0 &75.6 &76.3 &76.6 &76.8 &\cellcolor[HTML]{efefef}\textbf{77.3} &77.1 \\
Std. Dev. &\textbf{0.15} &0.22 &0.31 &0.36 &0.28 &\cellcolor[HTML]{efefef}0.22 &0.39 \\
\bottomrule
\end{tabular}
        \vspace{-2.5mm}
        \caption{\textbf{Patch size.} Leveraging the inherent simplicity and efficiency of our approach, we expand the receptive field of attention well beyond the conventional scope, surpassing sizes used in previous works such as PTv2~\cite{wu2022point}, which adopts a size of 16, and OctFormer~\cite{Wang2023OctFormer}, which uses 24.}\label{tab:ablation_patch_size} 
        \vspace{-5mm}
    \end{minipage}
\end{table}

\mypara{Serialization patterns.}
In \tabref{tab:ablation_serialization_interaction}, we explore the impact of various combinations of serialization patterns. Our experiments demonstrate that mixtures incorporating a broader range of patterns yield superior results when integrated with our Shuffle Order strategies. Furthermore, the additional computational overhead from introducing more serialization patterns is negligible. 
It is observed that relying on a single Shift Order cannot completely harness the potential offered by the four serialization patterns.

\mypara{Patch interaction.}
In \tabref{tab:ablation_serialization_interaction}, we also assess the effectiveness of each alternative patch interaction design. The default setting enables Shift Order, but the first row represents the baseline scenario using a single serialization pattern, indicative of the vanilla configurations of Shift Patch and Shift Dilation (one single serialization order is not shiftable). The results indicate that while Shift Patch and Shift Dilation are indeed effective, their latency is somewhat hindered by the dependency on attention masks, which compromises efficiency. Conversely, Shift Code, which utilizes multiple serialization patterns, offers a simple and efficient alternative that achieves comparable results to these traditional methods. Notably, when combined with Shuffle Order and all four serialization patterns, our strategy not only shows further improvement but also retains its efficiency.

\begin{table}[t]
    \begin{minipage}{0.48\textwidth}
    \centering
        \tablestyle{2.9pt}{1.08}
        \begin{tabular}{lx{8mm}x{8mm}x{9mm}x{9mm}x{8mm}x{8mm}}
\toprule
Indoor Sem. Seg.&\multicolumn{2}{c}{ScanNet~\cite{dai2017scannet}} &\multicolumn{2}{c}{ScanNet200~\cite{rozenberszki2022language}} &\multicolumn{2}{c}{S3DIS~\cite{armeni2016s3dis}} \\\cmidrule(lr){2-3} \cmidrule(lr){4-5} \cmidrule(lr){6-7}
Methods &Val &Test &Val &Test &Area5 &6-fold \\\midrule
\scratch MinkUNet~\cite{choy20194d} &72.2 &73.6 &25.0 &25.3 &65.4 &65.4 \\
\scratch ST~\cite{lai2022stratified} &74.3 &73.7 &- &- &72.0 &- \\
\scratch PointNeXt~\cite{qian2022pointnext} &71.5 &71.2 &- &- &70.5 &74.9 \\
\scratch OctFormer~\cite{Wang2023OctFormer} &75.7 &76.6 &32.6 &32.6 &- &- \\
\scratch Swin3D~\cite{yang2023swin3d} &76.4 &- &- &- &72.5 &76.9 \\
\scratch PTv1~\cite{zhao2021point} &70.6 &- &27.8 &- &70.4 &65.4 \\
\scratch PTv2~\cite{wu2022point} &75.4 &74.2 &30.2 &- &71.6 &73.5 \\
\cellcolor[HTML]{efefef}\scratch PTv3~(Ours) &\cellcolor[HTML]{efefef}77.5 &\cellcolor[HTML]{efefef}77.9 &\cellcolor[HTML]{efefef}35.2 &\cellcolor[HTML]{efefef}37.8 &\cellcolor[HTML]{efefef}73.4 &\cellcolor[HTML]{efefef}77.7 \\
\cellcolor[HTML]{efefef}\pretrain PTv3~(Ours) &\cellcolor[HTML]{efefef}\textbf{78.6} &\cellcolor[HTML]{efefef}\textbf{79.4}&\cellcolor[HTML]{efefef}\textbf{36.0} &\cellcolor[HTML]{efefef}\textbf{39.3}&\cellcolor[HTML]{efefef}\textbf{74.7} &\cellcolor[HTML]{efefef}\textbf{80.8} \\
\bottomrule
\end{tabular}

        \vspace{-3mm}
        \caption{\textbf{Indoor semantic segmentation.}}\label{tab:indoor_sem_seg}
        \vspace{2mm}
    \end{minipage} \\
    \begin{minipage}{0.48\textwidth}
    \centering
        \tablestyle{3pt}{1.08}
        \begin{tabular}{lrrrrrrrrr}\toprule
Method &Metric &Area1 &Area2 &Area3 &Area4 &Area5 &Area6 &6-Fold \\\midrule
\multirow{3}{*}{\scratch PTv2} &allAcc &92.30 &86.00 &92.98 &89.23 &91.24 &94.26 &90.76 \\
&mACC &88.44 &72.81 &88.41 &82.50 &77.85 &92.44 &83.13 \\
&mIoU &81.14 &61.25 &81.65 &69.06 &72.02 &85.95 &75.17 \\\midrule
\multirow{3}{*}{\scratch PTv3} &allAcc &93.22 &86.26 &94.56 &90.72 &91.67 &94.98 &91.53 \\
&mACC &89.92 &74.44 &94.45 &81.11 &78.92 &93.55 &85.31 \\
&\cellcolor[HTML]{efefef}mIoU &\cellcolor[HTML]{efefef}83.01 &\cellcolor[HTML]{efefef}63.42 &\cellcolor[HTML]{efefef}86.66 &\cellcolor[HTML]{efefef}71.34 &\cellcolor[HTML]{efefef}73.43 &\cellcolor[HTML]{efefef}87.31 &\cellcolor[HTML]{efefef}77.70 \\\midrule
\multirow{3}{*}{\pretrain PTv3} &allAcc &93.70 &90.34 &94.72 &91.87 &91.96 &94.98 &92.59 \\
&mACC &90.70 &78.40 &94.27 &86.61 &80.14 &93.80 &87.69 \\
&\cellcolor[HTML]{efefef}mIoU &\cellcolor[HTML]{efefef}\textbf{83.88} &\cellcolor[HTML]{efefef}\textbf{70.11} &\cellcolor[HTML]{efefef}\textbf{87.40} &\cellcolor[HTML]{efefef}\textbf{75.53} &\cellcolor[HTML]{efefef}\textbf{74.33} &\cellcolor[HTML]{efefef}\textbf{88.74} &\cellcolor[HTML]{efefef}\textbf{80.81} \\
\bottomrule
\end{tabular}
        \vspace{-3mm}
        \caption{\textbf{S3DIS 6-fold cross-validation.}}\label{tab:s3dis_6fold}
        \vspace{-4mm}
    \end{minipage}
\end{table}

\mypara{Positional encoding.}
In \tabref{tab:ablation_positional_encoding}, we benchmark our proposed CPE+ against conventional positional encoding, such as APE and RPE, as well as recent advanced solutions like cRPE and CPE. The results confirm that while RPE and cRPE are significantly more effective than APE, they also exhibit the inefficiencies previously discussed. Conversely, CPE and CPE+ emerge as superior alternatives. Although CPE+ incorporates slightly more parameters than CPE, it does not compromise our method's efficiency too much. Since CPEs operate prior to the attention phase rather than during it, they benefit from optimization like flash attention~\cite{dao2022flashattention, dao2023flashattention2}, which can be advantageous for our PTv3.

\mypara{Patch size.}
In \tabref{tab:ablation_patch_size}, we explore the scaling of the receptive field of attention by adjusting patch size. Beginning with a patch size of 16, a standard in prior point transformers, we observe that increasing the patch size significantly enhances performance. Moreover, as indicated in \tabref{tab:outdoor_model_efficiency} (benchmarked with NuScenes dataset), benefits from optimization techniques such as flash attention~\cite{dao2022flashattention, dao2023flashattention2}, the speed and memory efficiency are effectively managed.

\begin{table}[t]
    \begin{minipage}{0.48\textwidth}
    \centering
        \tablestyle{4.1pt}{1.08}
        \begin{tabular}{lccccccc}\toprule
Outdoor Sem. Seg. &\multicolumn{2}{c}{nuScenes~\cite{caesar2020nuscenes}} &\multicolumn{2}{c}{Sem.KITTI~\cite{behley2019semantickitti}} &\multicolumn{2}{c}{Waymo Val~\cite{sun2020waymo}} \\\cmidrule(lr){2-3} \cmidrule(lr){4-5} \cmidrule(lr){6-7}
Methods &Val &Test &Val &Test &mIoU &mAcc \\\midrule
\scratch MinkUNet~\cite{choy20194d} &73.3 &- &63.8 &- &65.9 &76.6 \\
\scratch SPVNAS~\cite{tang2020spvnas} &77.4 &- &64.7 &66.4 &- &- \\
\scratch Cylinder3D~\cite{zhu2021cylindrical} &76.1 &77.2 &64.3 &67.8 &- &- \\
\scratch AF2S3Net~\cite{Cheng2021af2s3net} &62.2 &78.0 &74.2 &70.8 &- &- \\
\scratch 2DPASS~\cite{yan20222dpass} &- &80.8 &69.3 &72.9 &- &- \\
\scratch SphereFormer~\cite{lai2023spherical} &78.4 &81.9 &67.8 &74.8 &69.9 &- \\
\scratch PTv2~\cite{wu2022point} &80.2 &82.6 &70.3 &72.6 &70.6 &80.2 \\
\cellcolor[HTML]{efefef}\scratch PTv3~(Ours) &\cellcolor[HTML]{efefef}80.4 &\cellcolor[HTML]{efefef}82.7 &\cellcolor[HTML]{efefef}70.8 &\cellcolor[HTML]{efefef}74.2 &\cellcolor[HTML]{efefef}71.3 &\cellcolor[HTML]{efefef}80.5 \\
\cellcolor[HTML]{efefef}\pretrain PTv3~(Ours) &\cellcolor[HTML]{efefef}\textbf{81.2} &\cellcolor[HTML]{efefef}\textbf{83.0} &\cellcolor[HTML]{efefef}\textbf{72.3} &\cellcolor[HTML]{efefef}\textbf{75.5} &\cellcolor[HTML]{efefef}\textbf{72.1} &\cellcolor[HTML]{efefef}\textbf{81.3} \\
\bottomrule
\end{tabular}
        \vspace{-3mm}
        \caption{\textbf{Outdoor semantic segmentation.}}\label{tab:outdoor_sem_seg}
        \vspace{2mm}
    \end{minipage} \\
    \begin{minipage}{0.48\textwidth}
    \centering
        \tablestyle{3.78pt}{1.08}
        \begin{tabular}{lccccccc}\toprule
Indoor Ins. Seg.&\multicolumn{3}{c}{ScanNet~\cite{dai2017scannet}} &\multicolumn{3}{c}{ScanNet200~\cite{rozenberszki2022language}} \\\cmidrule(lr){2-4} \cmidrule(lr){5-7}
PointGroup~\cite{jiang2020pointgroup} &mAP$_{25}$ &mAP$_{50}$ &mAP &mAP$_{25}$ &mAP$_{50}$ &mAP \\\midrule
\scratch MinkUNet~\cite{choy20194d} &72.8 &56.9 &36.0 &32.2 &24.5 &15.8 \\
\scratch PTv2~\cite{wu2022point} &76.3 &60.0 &38.3 &39.6 &31.9 &21.4 \\
\cellcolor[HTML]{efefef}\scratch PTv3~(Ours) &\cellcolor[HTML]{efefef}77.5 &\cellcolor[HTML]{efefef}61.7 &\cellcolor[HTML]{efefef}40.9 &\cellcolor[HTML]{efefef}40.1 &\cellcolor[HTML]{efefef}33.2 &\cellcolor[HTML]{efefef}23.1 \\
\cellcolor[HTML]{efefef}\pretrain PTv3~(Ours) &\cellcolor[HTML]{efefef}\textbf{78.9} &\cellcolor[HTML]{efefef}\textbf{63.5} &\cellcolor[HTML]{efefef}\textbf{42.1} &\cellcolor[HTML]{efefef}\textbf{40.8} &\cellcolor[HTML]{efefef}\textbf{34.1} &\cellcolor[HTML]{efefef}\textbf{24.0} \\
\bottomrule
\end{tabular}
        \vspace{-3mm}
        \caption{\textbf{Indoor instance segmentation.}}\label{tab:indoor_ins_seg}
        \vspace{2mm}
    \end{minipage} \\
    \begin{minipage}{0.48\textwidth}
    \centering
        \tablestyle{3.4pt}{1.08}
        \begin{tabular}{lccccccccc}\toprule
Data Efficient~\cite{hou2021exploring} &\multicolumn{4}{c}{Limited Reconstruction} &\multicolumn{4}{c}{Limited Annotation} \\\cmidrule(lr){2-5} \cmidrule(lr){6-9}
Methods &1\% &5\% &10\% &20\% &20 &50 &100 &200 \\\midrule
\scratch MinkUNet~\cite{choy20194d} &26.0 &47.8 &56.7 &62.9 &41.9 &53.9 &62.2 &65.5 \\
\scratch PTv2~\cite{wu2022point} &24.8 &48.1 &59.8 &66.3 &58.4 &66.1 &70.3 &71.2 \\
\cellcolor[HTML]{efefef}\scratch PTv3~(Ours) &\cellcolor[HTML]{efefef}25.8 &\cellcolor[HTML]{efefef}48.9 &\cellcolor[HTML]{efefef}61.0 &\cellcolor[HTML]{efefef}67.0 &\cellcolor[HTML]{efefef}60.1 &\cellcolor[HTML]{efefef}67.9 &\cellcolor[HTML]{efefef}71.4 &\cellcolor[HTML]{efefef}72.7 \\
\cellcolor[HTML]{efefef}\pretrain PTv3~(Ours) &\cellcolor[HTML]{efefef}\textbf{31.3} &\cellcolor[HTML]{efefef}\textbf{52.6} &\cellcolor[HTML]{efefef}\textbf{63.3} &\cellcolor[HTML]{efefef}\textbf{68.2} &\cellcolor[HTML]{efefef}\textbf{62.4} &\cellcolor[HTML]{efefef}\textbf{69.1} &\cellcolor[HTML]{efefef}\textbf{74.3} &\cellcolor[HTML]{efefef}\textbf{75.5} \\
\bottomrule
\end{tabular}
        \vspace{-3mm}
        \caption{\textbf{Data efficiency.}}\label{tab:data_efficient}
        \vspace{-4mm}
    \end{minipage}
\end{table}

\begin{table}[t]
    \begin{minipage}{0.48\textwidth}
    \centering
        \tablestyle{0pt}{1.08}
        \begin{tabular}{y{23mm}c x{8mm} x{8mm} x{8mm} x{8mm} x{8mm} x{8mm} x{11mm}}\toprule
\multicolumn{2}{l}{Waymo Obj. Det.}  &\multicolumn{2}{c}{Vehicle L2} &\multicolumn{2}{c}{Pedestrian L2} &\multicolumn{2}{c}{Cyclist L2} &Mean L2 \\\cmidrule(lr){3-4} \cmidrule(lr){5-6} \cmidrule(lr){7-8}  \cmidrule(lr){9-9}
Methods &\# &mAP &APH &mAP &APH &mAP &APH & mAPH \\\midrule
PointPillars~\cite{lang2019pointpillars} &1 &63.6 &63.1 &62.8 &50.3 &61.9 &59.9 &57.8 \\
CenterPoint~\cite{yin2021center} &1 &66.7 &66.2 &68.3 &62.6 &68.7 &67.6 &65.5 \\
SST~\cite{fan2022embracing} &1 &64.8 &64.4 &71.7 &63.0 &68.0 &66.9 &64.8 \\
SST-Center~\cite{fan2022embracing} &1 &66.6 &66.2 &72.4 &65.0 &68.9 &67.6 &66.3 \\
VoxSet~\cite{he2022voxset} &1 &66.0 &65.6 &72.5 &65.4 &69.0 &67.7 &66.2 \\
PillarNet~\cite{shi2022pillarnet} &1 &70.4 &69.9 &71.6 &64.9 &67.8 &66.7 &67.2 \\
FlatFormer~\cite{liu2023flatformer} &1 &69.0 &68.6 &71.5 &65.3 &68.6 &67.5 &67.2 \\
\cellcolor[HTML]{efefef}PTv3~(Ours) &\cellcolor[HTML]{efefef}1 &\cellcolor[HTML]{efefef}\textbf{71.2} &\cellcolor[HTML]{efefef}\textbf{70.8} &\cellcolor[HTML]{efefef}\textbf{76.3} &\cellcolor[HTML]{efefef}\textbf{70.4} &\cellcolor[HTML]{efefef}\textbf{71.5} &\cellcolor[HTML]{efefef}\textbf{70.4} &\cellcolor[HTML]{efefef}\textbf{70.5} \\
\midrule
CenterPoint~\cite{yin2021center} &2 &67.7 &67.2 &71.0 &67.5 &71.5 &70.5 &68.4 \\
PillarNet~\cite{shi2022pillarnet} &2 &71.6 &71.6 &74.5 &71.4 &68.3 &67.5 &70.2 \\
FlatFormer~\cite{liu2023flatformer} &2 &70.8 &70.3 &73.8 &70.5 &73.6 &72.6 &71.2 \\
\cellcolor[HTML]{efefef}PTv3~(Ours) &\cellcolor[HTML]{efefef}2 &\cellcolor[HTML]{efefef}\textbf{72.5} &\cellcolor[HTML]{efefef}\textbf{72.1} &\cellcolor[HTML]{efefef}\textbf{77.6} &\cellcolor[HTML]{efefef}\textbf{74.5} &\cellcolor[HTML]{efefef}\textbf{71.0} &\cellcolor[HTML]{efefef}\textbf{70.1} &\cellcolor[HTML]{efefef}\textbf{72.2} \\
\midrule
CenterPoint++~\cite{yin2021center} &3 &71.8 &71.4 &73.5 &70.8 &73.7 &72.8 &71.6 \\
SST~\cite{fan2022embracing} &3 &66.5 &66.1 &76.2 &72.3 &73.6 &72.8 &70.4 \\
FlatFormer~\cite{liu2023flatformer} &3 &71.4 &71.0 &74.5 &71.3 &74.7 &73.7 &72.0 \\
\cellcolor[HTML]{efefef}PTv3~(Ours) &\cellcolor[HTML]{efefef}3 &\cellcolor[HTML]{efefef}\textbf{73.0} &\cellcolor[HTML]{efefef}\textbf{72.5} &\cellcolor[HTML]{efefef}\textbf{78.0} &\cellcolor[HTML]{efefef}\textbf{75.0} &\cellcolor[HTML]{efefef}\textbf{72.3} &\cellcolor[HTML]{efefef}\textbf{71.4} &\cellcolor[HTML]{efefef}\textbf{73.0} \\
\bottomrule
\end{tabular}

        \vspace{-3mm}
        \caption{\textbf{Waymo object detection.} The colume with head name ``\#'' denotes the number of input frames.}\label{tab:outdoor_obj_det}
        \vspace{-4mm}
    \end{minipage}
\end{table}

\subsection{Results Comparision}
\label{sec:results}

We benchmark the performance of PTv3 against previous SOTA backbones and present the\prettyul{highest} results obtained for each benchmark. In our tables, Marker \scratch refers to a model trained from scratch, and \pretrain refers to a model trained with multi-dataset joint training (PPT~\cite{wu2023ppt}). An exhaustive comparison with earlier works is available in the\prettyul{Appendix}.

\mypara{Indoor semantic segmentation.}
In \tabref{tab:indoor_sem_seg}, we showcase the validation and test performance of PTv3 on the ScanNet v2~\cite{dai2017scannet} and ScanNet200~\cite{rozenberszki2022language} benchmarks, along with the Area 5 and 6-fold cross-validation~\cite{qi2017pointnet} on S3DIS~\cite{armeni2016s3dis} (details see \tabref{tab:s3dis_6fold}). We report the mean Intersection over Union (mIoU) percentages and benchmark these results against previous backbones. Even without pre-training, PTv3 outperforms PTv2 by 3.7\% on the ScanNet test split and by 4.2\% on the S3DIS 6-fold CV. The advantage of PTv3 becomes even more pronounced when scaling up the model with multi-dataset joint training~\cite{wu2023ppt}, widening the margin to 5.2\% on ScanNet and 7.3\% on S3DIS.

\mypara{Outdoor semantic segmentation.}
In \tabref{tab:outdoor_sem_seg}, we detail the validation and test results of PTv3 for the nuScenes~\cite{caesar2020nuscenes, fong2022panoptic} and SemanticKITTI~\cite{behley2019semantickitti} benchmarks and also include the validation results for the Waymo benchmark~\cite{sun2020waymo}. Performance metrics are presented as mIoU percentages by default, with a comparison to prior models. PTv3 demonstrates enhanced performance over the recent state-of-the-art model, SphereFormer, with a 2.0\% improvement on nuScenes and a 3.0\% increase on SemanticKITTI, both in the validation context. When pre-trained, PTv3's lead extends to 2.8\% for nuScenes and 4.5\% for SemanticKITTI.

\mypara{Indoor instance segmentation.}
In \tabref{tab:indoor_ins_seg}, we present PTv3's validation results on the ScanNet v2~\cite{dai2017scannet} and ScanNet200~\cite{rozenberszki2022language} instance segmentation benchmarks. We present the performance metrics as mAP, mAP$_{25}$, and mAP$_{50}$ and compare them against several popular backbones. To ensure a fair comparison, we standardize the instance segmentation framework by employing PointGroup~\cite{jiang2020pointgroup} across all tests, varying only the backbone. Our experiments reveal that integrating PTv3 as a backbone significantly enhances PointGroup, yielding a 4.9\% increase in mAP over MinkUNet. Moreover, fine-tuning a PPT pre-trained PTv3 provides an additional gain of 1.2\% mAP.

\mypara{Indoor data efficient.} In \tabref{tab:data_efficient}, we evaluate the performance of PTv3 on the ScanNet data efficient~\cite{hou2021exploring} benchmark. This benchmark tests models under constrained conditions with limited percentages of available reconstructions (scenes) and restricted numbers of annotated points. Across various settings, from 5\% to 20\% of reconstructions and from 20 to 200 annotations, PTv3 demonstrates strong performance. Moreover, the application of pre-training technologies further boosts PTv3's performance across all tasks.

\mypara{Outdoor object detection.}
In \tabref{tab:outdoor_obj_det}, we benchmark PTv3 against leading single-stage 3D detectors on the Waymo Object Detection benchmark. All models are evaluated using either anchor-based or center-based detection heads~\cite{yan2018second, yin2021center}, with a separate comparison for varying numbers of input frames. Our PTv3, engaged with CenterPoint, consistently outperforms both sparse convolutional~\cite{shi2022pillarnet, yin2021center} and transformer-based~\cite{fan2022embracing, he2022voxset} detectors, achieving significant gains even when compared with the recent state-of-the-art, FlatFormer~\cite{liu2023flatformer}. Notably, PTv3 surpasses FlatFormer by 3.3\% with a single frame as input and maintains a superiority of 1.0\% in multi-frame settings.

\mypara{Model efficiency.} We evaluate model efficiency based on average latency and memory consumption across real-world datasets. Efficiency metrics are measured on a single RTX 4090, excluding the first iteration to ensure steady-state measurements. We compared our PTv3 with multiple previous SOTAs. Specifically, we use the NuScenes dataset to assess outdoor model efficiency (see \tabref{tab:outdoor_model_efficiency}) and the ScanNet dataset for indoor model efficiency (see \tabref{tab:indoor_model_efficiency}). Our results demonstrate that PTv3 not only exhibits the lowest latency across all tested scenarios but also maintains reasonable memory consumption.

\begin{table}[!t]
    \begin{minipage}{0.48\textwidth}
    \centering
        \tablestyle{4pt}{1.08}
        \begin{tabular}{l|rrrrrr}\toprule
\multicolumn{2}{l}{Indoor Efficiency (ScanNet)} &\multicolumn{2}{c}{Training} &\multicolumn{2}{c}{Inference} \\ \cmidrule(lr){3-4} \cmidrule(lr){5-6}
Methods &Params. &Latency &Memory &Latency &Memory \\
\midrule
MinkUNet~\cite{choy20194d} &37.9M &267ms &\textbf{4.9G} &90ms &\textbf{4.7G} \\
OctFormer~\cite{Wang2023OctFormer} &44.0M &264ms &12.9G &86ms &12.5G \\
Swin3D~\cite{yang2023swin3d} &71.1M &602ms &13.6G &456ms &8.8G \\
PTv2~\cite{wu2022point} &12.8M &312ms &13.4G &191ms &18.2G \\
\cellcolor[HTML]{efefef}PTv3~(ours) &\cellcolor[HTML]{efefef}46.2M &\cellcolor[HTML]{efefef}\textbf{151ms} &\cellcolor[HTML]{efefef}6.8G &\cellcolor[HTML]{efefef}\textbf{61ms} &\cellcolor[HTML]{efefef}5.2G \\
\bottomrule
\end{tabular}
        \vspace{-3mm}
        \caption{\textbf{Indoor model efficiency.} }\label{tab:indoor_model_efficiency}
        \vspace{-4mm}
    \end{minipage}
\end{table}

\section{Conclusion and Discussion}
\label{sec:conclusion}

This paper presents Point Transformer V3, a stride towards overcoming the traditional trade-offs between accuracy and efficiency in point cloud processing. Guided by a novel interpretation of the \textbf{scaling principle} in backbone design, we propose that model performance is more profoundly influenced by scale than by complex design intricacies. By prioritizing efficiency over the accuracy of less impactful mechanisms, we harness the power of scale, leading to enhanced performance. Simply put, by making the model \textbf{simpler} and \textbf{faster}, we enable it to become \textbf{stronger}.

We discuss \textit{limitations and broader impacts} as follows:
\begin{itemize}[leftmargin=4mm, itemsep=0mm, topsep=0mm, partopsep=0mm]
    \item \textit{Attention mechanisum.} In prioritizing efficiency, PTv3 reverts to utilizing dot-product attention, which has been well-optimized through engineering efforts. However, we do note a reduction in convergence speed and a limitation in further scaling depth compared to vector attention. This issue also observed in recent advancements in transformer technology~\cite{xiao2023streamingllm}, is attributed to ``attention sinks'' stemming from the dot-product and softmax operations. Consequently, our findings reinforce the need for continued exploration of attention mechanisms.
    \item \textit{Scaling parameters.} PTv3 transcends the existing trade-offs between accuracy and efficiency, paving the way for investigating 3D transformers at larger parameter scales within given computational resources. While this exploration remains a topic for future work, current point cloud transformers already demonstrate an over-capacity for existing tasks. We advocate for a combined approach that scales up both the model parameters and the scope of data and tasks (e.g., learning from all available data, multi-task frameworks, and multi-modality tasks). Such an integrated strategy could fully unlock the potential of scaling in 3D representation learning.
    \item \textit{Multiple modalities.} Point cloud serialization provides a robust methodology for transforming n-dimensional data into a structured 1D format, effectively preserving spatial proximity. This technique can similarly be applied to image data, enabling its conversion into a language-style 1D structure that PTv3 can efficiently encode. This capability opens new avenues for the development of multimodal models that bridge 2D and 3D spaces, fostering opportunities for large-scale, synergistic pre-training that integrates both image and point cloud data.
\end{itemize}

\vspace{-2mm}
\section*{Acknowledgements}
\vspace{-2mm}
This work is supported in part by the National Natural Science Foundation of China (NO. 622014840), the National Key R\&D Program of China (NO. 2022ZD0160101), HKU Startup Fund, and HKU Seed Fund for Basic Research.

\appendix
\section*{Appendix}
For a thorough understanding of our Point Transformer V3 (PTv3), we have compiled a detailed Appendix. The table of contents below offers a quick overview and will guide to specific sections of interest.

\hypersetup{linkbordercolor=black,linkcolor=black}

\setlength{\cftbeforesecskip}{0.5em}
\cftsetindents{section}{0em}{1.8em}
\cftsetindents{subsection}{1em}{2.5em}

\etoctoccontentsline{part}{Appendix}
\localtableofcontents
\hypersetup{linkbordercolor=red,linkcolor=red}

\section{Implementation Details}
Our implementation primarily utilizes Pointcept~\cite{pointcept2023}, a specialized codebase focusing on point cloud perception and representation learning. For tasks involving outdoor object detection, we employ OpenPCDet~\cite{openpcdet2020}, which is tailored for LiDAR-based 3D object detection. The specifics of our implementation are detailed in this section.

\begin{table}[!t]
    \begin{minipage}{0.48\textwidth}
    \centering
        \tablestyle{4pt}{1.08}
        \begin{tabular}{lclc}\toprule
\multicolumn{2}{c}{Scratch} &\multicolumn{2}{c}{Joint Training~\cite{wu2023ppt}} \\
\cmidrule(lr){1-2} \cmidrule(lr){3-4}
Config &Value &Config &Value \\\midrule
optimizer &AdamW &optimizer &AdamW \\
scheduler &Cosine &scheduler &Cosine \\
criteria &CrossEntropy~(1) &criteria &CrossEntropy~(1) \\
&Lovasz~\cite{berman2018lovasz}~(1) & &Lovasz~\cite{berman2018lovasz}~(1) \\
learning rate &5e-3 &learning rate &5e-3 \\
block lr scaler &0.1 &block lr scaler &0.1 \\
weight decay &5e-2 &weight decay &5e-2 \\
batch size &12 &batch size &24 \\
datasets &ScanNet / &datasets &ScanNet (2) \\
&S3DIS / & &S3DIS (1) \\
&Struct.3D & &Struct.3D (4) \\
warmup epochs &40 &warmup iters &6k \\
epochs &800 &iters &120k \\
\bottomrule
\end{tabular}

        \vspace{-3mm}
        \caption{\textbf{Indoor semantic segmentation settings.} }\label{tab:indoor_sem_seg_settings}
        \vspace{2mm}
    \end{minipage}
    \begin{minipage}{0.48\textwidth}
    \centering
        \tablestyle{4pt}{1.08}
        \begin{tabular}{lclc}\toprule
\multicolumn{2}{c}{Scratch} &\multicolumn{2}{c}{Joint Training~\cite{wu2023ppt}} \\
\cmidrule(lr){1-2} \cmidrule(lr){3-4}
Config &Value &Config &Value \\\midrule
optimizer &AdamW &optimizer &AdamW \\
scheduler &Cosine &scheduler &Cosine \\
criteria &CrossEntropy~(1) &criteria &CrossEntropy~(1) \\
&Lovasz~\cite{berman2018lovasz}~(1) & &Lovasz~\cite{berman2018lovasz}~(1) \\
learning rate &2e-3 &learning rate &2e-3 \\
block lr scaler &1e-1 &block lr scaler &1e-1 \\
weight decay &5e-3 &weight decay &5e-3 \\
batch size &12 &batch size &24 \\
datasets &NuScenes / &datasets &NuScenes~(1) \\
&Sem.KITTI / & &Sem.KITTI~(1) \\
&Waymo & &Waymo~(1) \\
warmup epochs &2 &warmup iters &9k \\
epochs &50 &iters &180k \\
\bottomrule
\end{tabular}

        \vspace{-3mm}
        \caption{\textbf{Outdoor semantic segmentation settings.} }\label{tab:outdoor_sem_seg_settings}
        \vspace{2mm}
    \end{minipage}
    \begin{minipage}{0.48\textwidth}
    \centering
        \tablestyle{4pt}{1.08}
        \begin{tabular}{lclc}\toprule
\multicolumn{2}{c}{Ins. Seg.} &\multicolumn{2}{c}{Obj. Det} \\
\cmidrule(lr){1-2} \cmidrule(lr){3-4}
Config &Value &Config &Value \\\midrule
framework &PointGroup~\cite{jiang2020pointgroup} &framework &CenterPoint~\cite{yin2021center} \\
optimizer &AdamW &optimizer &Adam \\
scheduler &Cosine &scheduler &Cosine \\
learning rate &5e-3 &learning rate &3e-3 \\
block lr scaler &1e-1 &block lr scaler &1e-1 \\
weight decay &5e-2 &weight decay &1e-2 \\
batch size &12 &batch size &16 \\
datasets &ScanNet &datasets &Waymo \\
warmup epochs &40 &warmup epochs &0 \\
epochs &800 &epochs &24 \\
\bottomrule
\end{tabular}

        \vspace{-3mm}
        \caption{\textbf{Other downstream tasks settings.} }\label{tab:other_downstream_tasks_settings}
        \vspace{-4mm}
    \end{minipage}
\end{table}

\subsection{Training Settings}
\begin{table}[!t]
    \begin{minipage}{0.48\textwidth}
    \centering
        \tablestyle{14pt}{1.08}
        \begin{tabular}{lx{40mm}}\toprule
Config &Value \\\midrule
serialization pattern &Z + TZ + H + TH \\
patch interaction &Shift Order + Shuffle Order \\
positional encoding &xCPE \\
embedding depth &2 \\
embedding channels &32 \\
encoder depth &[2, 2, 6, 2] \\
encoder channels &[64, 128, 256, 512] \\
encoder num heads &[4, 8, 16, 32] \\
encoder patch size &[1024, 1024, 1024, 1024] \\
decoder depth &[1, 1, 1, 1] \\
decoder channels &[64, 64, 128, 256] \\
decoder num heads &[4, 4, 8, 16] \\
decoder patch size &[1024, 1024, 1024, 1024] \\
down stride &[$\times$2, $\times$2, $\times$2, $\times$2] \\
mlp ratio &4 \\
qkv bias &True \\
drop path &0.3 \\
\bottomrule
\end{tabular}
        \vspace{-3mm}
        \caption{\textbf{Model settings.} }\label{tab:appendix_model_settings}
        \vspace{2mm}
    \end{minipage}
    \begin{minipage}{0.48\textwidth}
    \centering
        \tablestyle{1.5pt}{1.07}
        \begin{tabular}{llcc}\toprule
Augmentations &Parameters &Indoor &Outdoor \\\midrule
random dropout &dropout ratio: 0.2, p: 0.2 &\checkmark &- \\
random rotate &axis: z, angle: [-1, 1], p: 0.5 &\checkmark &\checkmark \\
&axis: x, angle: [-1 / 64, 1 / 64], p: 0.5 &\checkmark &- \\
&axis: y, angle: [-1 / 64, 1 / 64], p: 0.5 &\checkmark &- \\
random scale &scale: [0.9, 1.1] &\checkmark &\checkmark \\
random flip &p: 0.5 &\checkmark &\checkmark \\
random jitter &sigma: 0.005, clip: 0.02 &\checkmark &\checkmark \\
elastic distort & params: [[0.2, 0.4], [0.8, 1.6]] &\checkmark &- \\
auto contrast &p: 0.2 &\checkmark &- \\
color jitter &std: 0.05; p: 0.95 &\checkmark &- \\
grid sampling &grid size: 0.02 (indoor), 0.05 (outdoor) &\checkmark &\checkmark \\
sphere crop &ratio: 0.8, max points: 128000 &\checkmark &- \\
normalize color &p: 1 &\checkmark &- \\
\bottomrule
\end{tabular}
        \vspace{-3mm}
        \caption{\textbf{Data augmentations.} }\label{tab:appendix_data_augmentations}
        \vspace{-6mm}
    \end{minipage}
\end{table}
\mypara{Indoor semantic segmentation.} 
The settings for indoor semantic segmentation are outlined in \tabref{tab:indoor_sem_seg_settings}. The two leftmost columns describe the parameters for training from scratch using a single dataset. To our knowledge, this represents the first initiative to standardize training settings across different indoor benchmarks with a unified approach. The two rightmost columns describe the parameters for multi-dataset joint training~\cite{wu2023ppt} with PTv3, maintaining similar settings to the scratch training but with an increased batch size. The numbers in brackets indicate the relative weight assigned to each dataset (criteria) in the mix.

\mypara{Outdoor semantic segmentation.}
The configuration for outdoor semantic segmentation, presented in \tabref{tab:outdoor_sem_seg_settings}, follows a similar format to that of indoor. We also standardize the training settings across three outdoor datasets. Notably, PTv3 operates effectively without the need for point clipping within a specific range, a step that is typically essential in current models. Furthermore, we extend our methodology to multi-dataset joint training with PTv3, employing settings analogous to scratch training but with augmented batch size. The numbers in brackets represent the proportional weight assigned to each dataset in the training mix.

\mypara{Other Downstream Tasks.}
We outline our configurations for indoor instance segmentation and outdoor object detection in \tabref{tab:other_downstream_tasks_settings}. For indoor instance segmentation, we use PointGroup~\cite{jiang2020pointgroup} as our foundational framework, a popular choice in 3D representation learning~\cite{xie2020pointcontrast, hou2021exploring, wu2023masked, wu2023ppt}. Our configuration primarily follows PointContrast~\cite{xie2020pointcontrast}, with necessary adjustments made for PTv3 compatibility. Regarding outdoor object detection, we adhere to the settings detailed in FlatFormer~\cite{liu2023flatformer} and implement CenterPoint as our base framework to assess PTv3's effectiveness. It's important to note that PTv3 is versatile and can be integrated with various other frameworks due to its backbone nature.

\begin{table}[!t]
    \begin{minipage}{0.48\textwidth}
    \centering
        \tablestyle{13.5pt}{1.08}
        \begin{tabular}{l|cccc}\toprule
Block &BN &LN &BN &\cellcolor[HTML]{efefef}LN \\
Pooling &BN &LN &LN &\cellcolor[HTML]{efefef}BN \\\midrule
Perf. &76.7 &76.1 &75.6 &\cellcolor[HTML]{efefef}\textbf{77.3} \\
\bottomrule
\end{tabular}
        \vspace{-3mm}
        \caption{\textbf{Nomalization layer.} }\label{tab:appendix_ablation_normalization}
        \vspace{2mm}
    \end{minipage}
    \begin{minipage}{0.48\textwidth}
    \centering
        \tablestyle{11.5pt}{1.07}
        \begin{tabular}{l|ccc}\toprule
Block &Traditional &Post-Norm &\cellcolor[HTML]{efefef}Pre-Norm \\\midrule
Perf. &76.6 &72.3 &\cellcolor[HTML]{efefef}\textbf{77.3} \\
\bottomrule
\end{tabular}
        \vspace{-3mm}
        \caption{\textbf{Block structure.} }\label{tab:appendix_ablation_block}
        \vspace{-6mm}
    \end{minipage}
\end{table}

\subsection{Model Settings}
As briefly described in \secref{subsec:network_details}, here we delve into the detailed model configurations of our PTv3, which are comprehensively listed in \tabref{tab:appendix_model_settings}. This table serves as a blueprint for components within serialization-based point cloud transformers, encapsulating models like OctFormer~\cite{Wang2023OctFormer} and FlatFormer~\cite{liu2023flatformer} within the outlined frameworks, except for certain limitations discussed in \secref{sec:related}. Specifically, OctFormer can be interpreted as utilizing a single z-order serialization with patch interaction enabled by Shift Dilation. Conversely, FlatFormer can be characterized by its window-based serialization approach, facilitating patch interaction through Shift Order.

\vspace{-1mm}
\subsection{Data Augmentations}
\vspace{-1mm}
The specific configurations of data augmentations implemented for PTv3 are detailed in \tabref{tab:appendix_data_augmentations}. We unify augmentation pipelines for both indoor and outdoor scenarios separately, and the configurations are shared by all tasks within each domain. Notably, we observed that PTv3 does not depend on point clipping within a specific range, a process often crucial for existing models.

\begin{table}[!t]
    \begin{minipage}{0.48\textwidth}
    \centering
        \tablestyle{10pt}{1.08}
        \begin{tabular}{lccc}\toprule
Methods &Year &Val &Test \\\midrule
\scratch PointNet++~\cite{qi2017pointnet++} & 2017 & 53.5 & 55.7 \\
\scratch 3DMV~\cite{dai20183dmv} & 2018 & -& 48.4\\
\scratch PointCNN~\cite{li2018pointcnn} & 2018 & -& 45.8\\
\scratch SparseConvNet~\cite{graham20183d} & 2018 & 69.3 & 72.5 \\
\scratch PanopticFusion~\cite{narita2019panopticfusion} & 2019 & -& 52.9\\

\scratch PointConv~\cite{wu2019pointconv} & 2019 & 61.0 & 66.6\\
\scratch JointPointBased~\cite{chiang2019unified}& 2019 & 69.2 & 63.4\\
\scratch KPConv~\cite{thomas2019kpconv} & 2019 & 69.2 & 68.6 \\
\scratch PointASNL~\cite{yan2020pointasnl}& 2020 & 63.5 & 66.6\\
\scratch SegGCN~\cite{lei2020seggcn}& 2020 & -& 58.9\\
\scratch RandLA-Net~\cite{hu2020randla} & 2020 & - & 64.5 \\
\scratch JSENet~\cite{hu2020jsenet} & 2020 & - & 69.9 \\
\scratch FusionNet~\cite{zhang2020deep} & 2020 & - & 68.8 \\
\scratch FastPointTransformer~\cite{park2022fast} &2022 &72.4 &- \\
\scratch SratifiedTranformer~\cite{lai2022stratified} &2022 &74.3 &73.7 \\
\scratch PointNeXt~\cite{qian2022pointnext} &2022 &71.5 &71.2 \\
\scratch LargeKernel3D~\cite{chen2023largekernel3d} & 2023 &73.5 &73.9 \\
\scratch PointMetaBase~\cite{lin2023pointmetabase} & 2023 &72.8 &71.4 \\
\scratch PointConvFormer~\cite{wu2023pointconvformer} & 2023 &74.5 &74.9 \\
\scratch OctFormer~\cite{Wang2023OctFormer} &2023 &75.7 &76.6 \\
\scratch Swin3D~\cite{yang2023swin3d} &2023 &77.5 &77.9 \\
\pretrain\ + Supervised~\cite{yang2023swin3d} &2023 &76.7 &77.9 \\
\scratch MinkUNet~\cite{choy20194d} &2019 &72.2 &73.6 \\
\pretrain\ + PC~\cite{xie2020pointcontrast} &2020 &74.1 &- \\
\pretrain\ + CSC~\cite{hou2021exploring} &2021 &73.8 &- \\
\pretrain\ + MSC~\cite{wu2023masked} &2023 &75.5 &- \\
\pretrain\ + GC~\cite{wang2024gc} &2024 &75.7 &- \\
\pretrain\ + PPT~\cite{wu2023ppt} &2024 &76.4 &76.6 \\
\scratch OA-CNNs~\cite{peng2024oacnns} &2024 &76.1 &75.6 \\
\scratch PTv1~\cite{zhao2021point} &2021 &70.6 &- \\
\scratch PTv2~\cite{wu2022point} &2022 &75.4 &74.2 \\
\cellcolor[HTML]{efefef}\scratch PTv3~(Ours) &\cellcolor[HTML]{efefef}2024 &\cellcolor[HTML]{efefef}77.5 &\cellcolor[HTML]{efefef}77.9 \\
\cellcolor[HTML]{efefef}\pretrain\ + PPT~\cite{wu2023ppt} &\cellcolor[HTML]{efefef}2024 &\cellcolor[HTML]{efefef}\textbf{78.6} &\cellcolor[HTML]{efefef}\textbf{79.4} \\
\bottomrule
\end{tabular}
        \vspace{-3mm}
        \caption{\textbf{ScanNet V2 semantic segmentation.} }\label{tab:appendix_scannet}
        \vspace{-6mm}
    \end{minipage}
\end{table}

\section{Additional Ablations}
In this section, we present further ablation studies focusing on macro designs of PTv3, previously discussed in \secref{subsec:network_details}.

\begin{table}[!t]
    \begin{minipage}{0.48\textwidth}
    \centering
        \tablestyle{10pt}{1.08}
        \begin{tabular}{lccc}\toprule
Methods &Year &Area5 &6-fold \\\midrule
\scratch PointNet~\cite{qi2017pointnet} &2017 &41.1 &47.6\\
\scratch SegCloud~\cite{tchapmi2017segcloud} &2017 &48.9 &-\\
\scratch TanConv~\cite{tatarchenko2018tangent} &2018 &52.6 &-\\
\scratch PointCNN~\cite{li2018pointcnn} &2018 &57.3 &65.4\\
\scratch ParamConv~\cite{wang2018deep} &2018 &58.3 &-\\
\scratch PointWeb~\cite{zhao2019pointweb} &2019 &60.3 &66.7\\
\scratch HPEIN~\cite{jiang2019hierarchical} &2019 &61.9 &-\\
\scratch KPConv~\cite{thomas2019kpconv} &2019 &67.1 &70.6\\
\scratch GACNet~\cite{wang2019graph} &2019 &62.9 &-\\
\scratch PAT~\cite{yang2019modeling} &2019 &60.1 &-\\
\scratch SPGraph~\cite{landrieu2018large} &2018 &58.0 &62.1\\
\scratch SegGCN~\cite{lei2020seggcn} &2020 &63.6 &-\\
\scratch PAConv~\cite{xu2021paconv} &2021 &66.6 &-\\
\scratch StratifiedTransformer~\cite{lai2022stratified} &2022 &72.0 &- \\
\scratch PointNeXt~\cite{qian2022pointnext} &2022 &70.5 &74.9 \\
\scratch SuperpointTransformer~\cite{robert2023spt} & 2023 &68.9 &76.0 \\
\scratch PointMetaBase~\cite{lin2023pointmetabase} & 2023 &72.0 &77.0 \\
\scratch Swin3D~\cite{yang2023swin3d} &2023 &72.5 &76.9 \\
\pretrain\ + Supervised~\cite{yang2023swin3d} &2023 &74.5 &79.8 \\
\scratch MinkUNet~\cite{choy20194d} &2019 &65.4 &65.4 \\
\pretrain\ + PC~\cite{xie2020pointcontrast} &2020 &70.3 &- \\
\pretrain\ + CSC~\cite{hou2021exploring} &2021 &72.2 &- \\
\pretrain\ + MSC~\cite{wu2023masked} &2023 &70.1 &- \\
\pretrain\ + GC~\cite{wang2024gc} &2024 &72.0 &- \\
\pretrain\ + PPT~\cite{wu2023ppt} &2024 &72.7 &78.1 \\
\scratch PTv1~\cite{zhao2021point} &2021 &70.4 &65.4 \\
\scratch PTv2~\cite{wu2022point} &2022 &71.6 &73.5 \\
\cellcolor[HTML]{efefef}\scratch PTv3~(Ours) &\cellcolor[HTML]{efefef}2024 &\cellcolor[HTML]{efefef}73.4 &\cellcolor[HTML]{efefef}77.7 \\
\cellcolor[HTML]{efefef}\pretrain\ + PPT~\cite{wu2023ppt} &\cellcolor[HTML]{efefef}2024 &\cellcolor[HTML]{efefef}\textbf{74.7} &\cellcolor[HTML]{efefef}\textbf{80.8} \\
\bottomrule
\end{tabular}
        \vspace{-3mm}
        \caption{\textbf{S3DIS semantic segmentation.} }\label{tab:appendix_s3dis}
        \vspace{-6mm}
    \end{minipage}
\end{table}

\subsection{Nomalization Layer}
Previous point transformers employ Batch Normalization (BN), which can lead to performance variability depending on the batch size. This variability becomes particularly problematic in scenarios with memory constraints that require small batch sizes or in tasks demanding dynamic or varying batch sizes. To address this issue, we have gradually transitioned to Layer Normalization (LN). Our final, empirically determined choice is to implement Layer Normalization in the attention blocks while retaining Batch Normalization in the pooling layers (see \tabref{tab:appendix_ablation_normalization}).

\subsection{Block Structure}
Previous point transformers use a traditional block structure that sequentially applies an operator, a normalization layer, and an activation function. While effective, this approach can sometimes complicate training deeper models due to issues like vanishing gradients or the need for careful initialization and learning rate adjustments~\cite{xiong2020layer}. Consequently, we explored adopting a more modern block structure, such as pre-norm and post-norm. The pre-norm structure, where a normalization layer precedes the operator, can stabilize training by ensuring normalized inputs for each layer~\cite{child2019generating}. In contrast, the post-norm structure places a normalization layer right after the operator, potentially leading to faster convergence but with less stability~\cite{vaswani2017attention}. Our experimental results (see \tabref{tab:appendix_ablation_block}) indicate that the pre-norm structure is more suitable for our PTv3, aligning with findings in recent transformer-based models~\cite{xiong2020layer}.

\section{Additional Comparision}
In this section, we expand upon the combined results table for semantic segmentation (\tabref{tab:indoor_sem_seg} and \tabref{tab:outdoor_sem_seg}) from our main paper, offering a more detailed breakdown of results alongside the respective publication years of previous works. This comprehensive result table is designed to assist readers in tracking the progression of research efforts in 3D representation learning. Marker \scratch refers to the result from a model trained from scratch, and \pretrain refers to the result from a pre-trained model.

\subsection{Indoor Semantic Segmentation}

We conduct a detailed comparison of pre-training technologies and backbones on the ScanNet v2~\cite{dai2017scannet} (see \tabref{tab:appendix_scannet}) and S3DIS~\cite{armeni2016s3dis} (see \tabref{tab:appendix_s3dis}) datasets. ScanNet v2 comprises 1,513 room scans reconstructed from RGB-D frames, divided into 1,201 training scenes and 312 for validation. In this dataset, model input point clouds are sampled from the vertices of reconstructed meshes, with each point assigned a semantic label from 20 categories (e.g., wall, floor, table). The S3DIS dataset for semantic scene parsing includes 271 rooms across six areas from three buildings. Following a common practice~\cite{tchapmi2017segcloud,qi2017pointnet++,zhao2021point}, we withhold area 5 for testing and perform a 6-fold cross-validation. Different from ScanNet v2, S3DIS densely sampled points on mesh surfaces, annotated into 13 categories. Consistent with standard practice~\citep{qi2017pointnet++}. We employ the mean class-wise intersection over union (mIoU) as the primary evaluation metric for indoor semantic segmentation.

\begin{table}[!t]
    \begin{minipage}{0.48\textwidth}
    \centering
        \tablestyle{15pt}{1.08}
        \begin{tabular}{lccc}\toprule
Methods &Year &Val &Test \\\midrule
\scratch SPVNAS~\cite{tang2020spvnas} &2020 &64.7 &66.4 \\
\scratch Cylinder3D~\cite{zhu2021cylindrical} &2021 &64.3 &67.8 \\
\scratch PVKD~\cite{hou2022pvkd} &2022 &- &71.2 \\
\scratch 2DPASS~\cite{yan20222dpass} &2022 &69.3 &72.9 \\
\scratch WaffleIron~\cite{puy23waffleiron} &2023 &68.0 &70.8 \\
\scratch SphereFormer~\cite{lai2023spherical} &2023 &67.8 &74.8 \\
\scratch RangeFormer~\cite{kong2023rangeformer} &2023 &67.6 &73.3 \\
\scratch MinkUNet~\cite{choy20194d} &2019 &63.8 &- \\
\pretrain\ + M3Net~\cite{liu2024m3net} &2024 &69.9 &- \\
\pretrain\ + PPT~\cite{wu2023ppt} &2024 &71.4 &- \\
\scratch OA-CNNs~\cite{peng2024oacnns} &2024 &70.6 &- \\
\scratch PTv2~\cite{wu2022point} &2022 &70.3 &72.6 \\
\cellcolor[HTML]{efefef}\scratch PTv3~(Ours) &\cellcolor[HTML]{efefef}2024 &\cellcolor[HTML]{efefef}70.8 &\cellcolor[HTML]{efefef}74.2 \\
\cellcolor[HTML]{efefef}\pretrain\ + M3Net~\cite{liu2024m3net} &\cellcolor[HTML]{efefef}2024 &\cellcolor[HTML]{efefef}72.0 &\cellcolor[HTML]{efefef}75.1 \\
\cellcolor[HTML]{efefef}\pretrain\ + PPT~\cite{wu2023ppt} &\cellcolor[HTML]{efefef}2024 &\cellcolor[HTML]{efefef}\textbf{72.3} &\cellcolor[HTML]{efefef}\textbf{75.5} \\
\bottomrule
\end{tabular}
        \vspace{-3mm}
        \caption{\textbf{SemanticKITTI semantic segmentation.} }\label{tab:appendix_semantic_kitti}
        \vspace{2mm}
    \end{minipage}
    \begin{minipage}{0.48\textwidth}
    \centering
        \tablestyle{15pt}{1.08}
        \begin{tabular}{lrrrr}\toprule
Methods &Year &Val &Test \\\midrule
\scratch SPVNAS~\cite{tang2020spvnas} &2020 &77.4 &- \\
\scratch Cylinder3D~\cite{zhu2021cylindrical} &2021 &76.1 &77.2 \\
\scratch PVKD~\cite{hou2022pvkd} &2022 &- &76.0 \\
\scratch 2DPASS~\cite{yan20222dpass} &2022 &- &80.8 \\
\scratch SphereFormer~\cite{lai2023spherical} &2023 &78.4 &81.9 \\
\scratch RangeFormer~\cite{kong2023rangeformer} &2023 &78.1 &80.1 \\
\scratch MinkUNet~\cite{choy20194d} &2019 &73.3 &- \\
\pretrain\ + M3Net~\cite{liu2024m3net} &2024 &79.0 &- \\
\pretrain\ + PPT~\cite{wu2023ppt} &2024 &78.6 &\textbf{-} \\
\scratch OA-CNNs~\cite{peng2024oacnns} &2024 &78.9 &- \\
\scratch PTv2~\cite{wu2022point} &2022 &80.2 &82.6 \\
\cellcolor[HTML]{efefef}\scratch PTv3~(Ours) &\cellcolor[HTML]{efefef}2024 &\cellcolor[HTML]{efefef}80.4 &\cellcolor[HTML]{efefef}82.7 \\
\cellcolor[HTML]{efefef}\pretrain\ + M3Net~\cite{liu2024m3net} &\cellcolor[HTML]{efefef}2024 &\cellcolor[HTML]{efefef}80.9 &\cellcolor[HTML]{efefef}\textbf{83.1} \\
\cellcolor[HTML]{efefef}\pretrain\ + PPT~\cite{wu2023ppt} &\cellcolor[HTML]{efefef}2024 &\cellcolor[HTML]{efefef}\textbf{81.2} &\cellcolor[HTML]{efefef}83.0 \\
\bottomrule
\end{tabular}
        \vspace{-3mm}
        \caption{\textbf{NuScenes semantic segmentation.} }\label{tab:appendix_nuscenes}
        \vspace{-6mm}
    \end{minipage}
\end{table}

\subsection{Outdoor Semantic Segmentation}
We extend our comprehensive evaluation of pre-training technologies and backbones to outdoor semantic segmentation tasks, focusing on the SemanticKITTI~\cite{behley2019semantickitti}(see \tabref{tab:appendix_semantic_kitti}) and NuScenes~\cite{caesar2020nuscenes} (see \tabref{tab:appendix_nuscenes}) datasets. SemanticKITTI is derived from the KITTI Vision Benchmark Suite and consists of 22 sequences, with 19 for training and the remaining 3 for testing. It features richly annotated LiDAR scans, offering a diverse array of driving scenarios. Each point in this dataset is labeled with one of 28 semantic classes, encompassing various elements of urban driving environments. NuScenes, on the other hand, provides a large-scale dataset for autonomous driving, comprising 1,000 diverse urban driving scenes from Boston and Singapore. For outdoor semantic segmentation, we also employ the mean class-wise intersection over union (mIoU) as the primary evaluation metric for outdoor semantic segmentation.

{
\small
\bibliographystyle{ieeenat_fullname}
\bibliography{main}
}

\end{document}